\documentclass{article}

\PassOptionsToPackage{numbers,sort&compress}{natbib}

\usepackage[final]{neurips_2025}  

\usepackage[utf8]{inputenc}
\usepackage[T1]{fontenc}
\usepackage{microtype}
\usepackage{minitoc}
\usepackage{titletoc}

\usepackage{booktabs}
\usepackage{multirow}
\usepackage{siunitx}

\usepackage{ragged2e}

\usepackage{amsmath,amssymb,amsfonts,amsthm}
\usepackage{nicefrac}
\usepackage{bm}

\usepackage{graphicx}
\usepackage{subcaption}
\usepackage{booktabs}
\usepackage{array,longtable}
\usepackage[table]{xcolor}
\newcolumntype{R}{>{\raggedright\arraybackslash}p{.18\textwidth}}

\usepackage{enumitem}

\usepackage{algorithm}
\usepackage{algpseudocode}

\usepackage{listings}
\usepackage{multirow}
\usepackage{multicol}
\definecolor{codegreen}{rgb}{0,0.6,0}
\definecolor{codegray} {rgb}{0.5,0.5,0.5}
\definecolor{codepurple}{rgb}{0.58,0,0.82}
\definecolor{codeblue} {rgb}{0,0,0.6}
\definecolor{backcolour}{rgb}{0.96,0.96,0.94}
\lstdefinestyle{mystyle}{
  backgroundcolor=\color{backcolour},
  commentstyle=\color{codegreen}\itshape,
  keywordstyle=\color{codeblue}\bfseries,
  numberstyle=\tiny\color{codegray},
  stringstyle=\color{codepurple},
  basicstyle=\ttfamily\small,
  breaklines=true,
  frame=tb,
  numbers=left,
  numbersep=5pt,
}
\usepackage{CJKutf8}

\usepackage[colorlinks,linkcolor=blue,citecolor=green,urlcolor=cyan]{hyperref}
\usepackage[capitalise,noabbrev]{cleveref}

\newtheorem{proposition}{Proposition}[section]

\title{Geo-Sign: Hyperbolic Contrastive Regularisation for Geometrically Aware Sign Language Translation}
\author{%
  Edward Fish\\
  CVSSP, University of Surrey\\
  \texttt{edward.fish@surrey.ac.uk}
  \and
  \textbf{Richard Bowden}\\
  CVSSP, University of Surrey\\
  \texttt{r.bowden@surrey.ac.uk}
}

\begin{document}
\maketitle

\begin{abstract}
Recent progress in Sign Language Translation (SLT) has focussed primarily on improving the representational capacity of large language models to incorporate Sign Language features. This work explores an alternative direction: enhancing the geometric properties of skeletal representations themselves. We propose Geo-Sign, a method that leverages the properties of hyperbolic geometry to model the hierarchical structure inherent in sign language kinematics. By projecting skeletal features derived from Spatio-Temporal Graph Convolutional Networks (ST-GCNs) into the Poincaré ball model, we aim to create more discriminative embeddings, particularly for fine-grained motions like finger articulations. We introduce a hyperbolic projection layer, a weighted Fréchet mean aggregation scheme, and a geometric contrastive loss operating directly in hyperbolic space. These components are integrated into an end-to-end translation framework as a regularisation function, to enhance the representations within the language model. This work demonstrates the potential of hyperbolic geometry to improve skeletal representations for Sign Language Translation, improving on SOTA RGB methods while preserving privacy and improving computational efficiency. Code available here: \href{https://github.com/ed-fish/Geo-Sign}{https://github.com/ed-fish/geo-sign}.
\end{abstract}

\section{Introduction}
\label{sec:introduction}
Sign Languages are rich, multi-channel linguistic systems where meaning is conveyed through a composition of movements involving the upper body, hands, face, and mouth. Automatic Sign Language Translation (SLT) is an established research area focused on developing methods to convert these visual expressions directly into text. While Sign Languages are expressed via fluid multi-articulator kinematics, a persistent challenge for SLT methods lies in creating feature representations that concurrently preserve fine-grained, local details (e.g., subtle finger configurations) while embedding the global structure inherent in larger, overarching body motions. Effectively modelling these multi-scale and relational dynamics within a suitable geometric embedding space remains a central hurdle.

Spatio-Temporal Graph Convolutional Networks (ST-GCNs) offer a natural way to encode these hierarchical relationships by treating the body’s joints and bones as nodes and edges in a graph \cite{yan2018spatial}. However, when their learned representations are projected into standard Euclidean geometry for processing via a Large Language Model (LLM), essential fine-grained relational distances and movements can become blurred. For instance, the sign for “\emph{water}” in American Sign Language (ASL) is communicated by forming a W shape with the fingers and tapping the chin twice (a fine-grained, "leaf-level" articulation), immediately followed by a sweeping hand movement away from the body (a "branch-level" gesture). When these features are aggregated in Euclidean geometry, the large translation and rotation of the wrist could dominate the vector’s norm, effectively “pulling” the embedding toward the global motion and compressing the subtle finger tap into a vanishing tail. Consequently, two signs that differ only in the timing or precision of that tap, which may be critical to lexical meaning, can become nearly indistinguishable once projected into flat Euclidean space.

Large vision-based models \cite{hu2021signbert, hu2023signbert+,guo-etal-2024-unsupervised} appear to be able to implicitly learn these hierarchical structures through extensive video pre-training and visual inductive biases. However, they do so at significant computational cost and with privacy concerns, as they retain identifiable facial and background details that skeletal representations inherently discard.
\begin{figure}[t]
    \centering
    \includegraphics[width=1\linewidth]{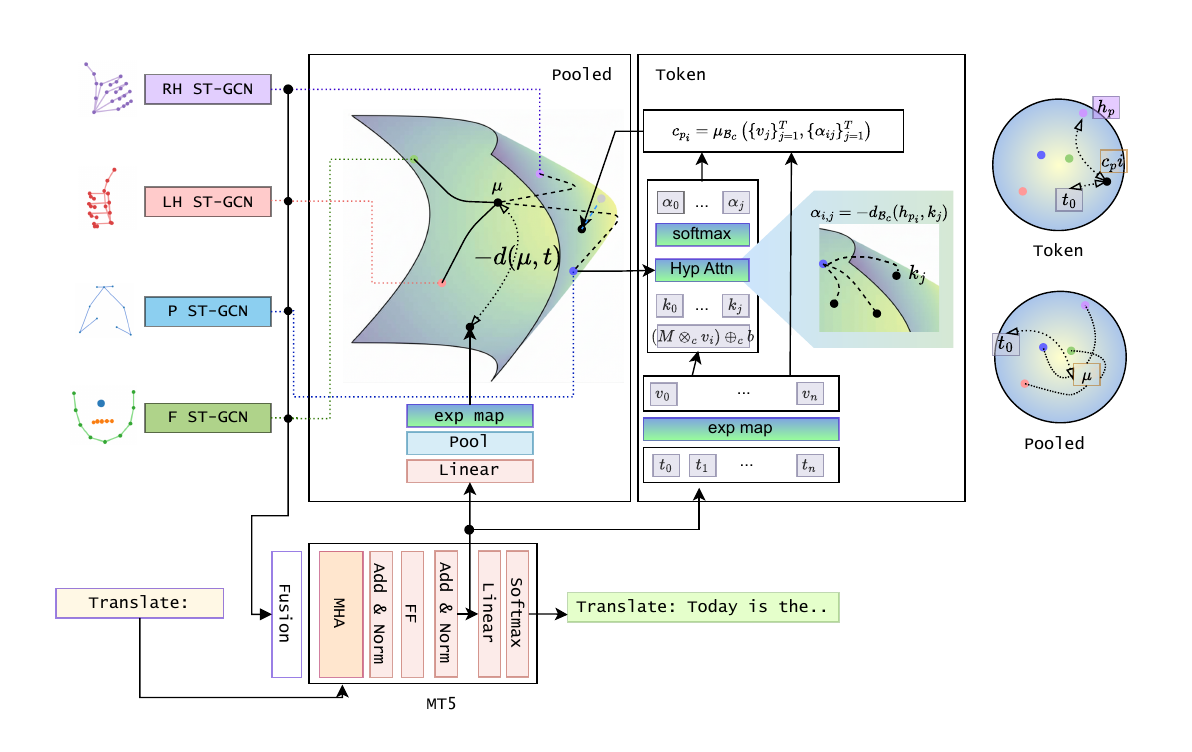} 
        \caption{Geo-Sign’s hyperbolic framework: (\textbf{Left}) Skeletal features from ST-GCN's for different body parts are projected into a Poincaré ball whose curvature is learned, while the original branch fuses the features for processing via the MT5 language model. (\textbf{Pooled}) The pose features are aggregated via Frechet Mean in Eq.\ref{alg:frechet_mean_balanced_v1}, while the text embeddings from the final layer of the MT5 model are pooled and projected to the hyperbolic manifold. Geodesic distance between the text embedding and the mean pose features are minimised for positive samples using the contrastive loss in Eq.\ref{eq:method_hyperbolic_contrastive_loss_overall_balanced_v1}. (\textbf{Token}) Alternatively, hyperbolic pose features are used as attention queries against all text embeddings to generate a pose-contextual text embedding. Note the movement of the text features $c_{pi}$ in grey towards the pose feature in blue. (\textbf{Right}) A representation of the Poincaré disk demonstrating the difference between Token, and Pooled methods in the tangent space. }\label{fig:hyperbolic_intro} 
\end{figure}

This work introduces hyperbolic geometry as a means to fundamentally enhance skeletal representations for SLT. Unlike Euclidean space, where volume grows polynomially with radius and can flatten hierarchical structures, hyperbolic manifolds exhibit exponential volume growth. This property is naturally suited to encoding the compositional, tree-like structures found in sign language kinematics. As illustrated in \Cref{fig:hyperbolic_intro}, in the Poincaré ball model $\mathbb{B}_{c}^{d_{\text{hyp}}}$ (with curvature $\kappa = -c < 0$), distances between points near the boundary expand exponentially relative to their Euclidean separation. This provides ample "space" to distinguish nuanced motions (e.g., an open versus a closed fist), while regions near the origin behave more like Euclidean space, suitable for representing broader phrase-level semantics. A key aspect of our approach is that we learn the curvature parameter $c$ end-to-end via Riemannian optimization. This allows the manifold to dynamically adapt its "zoom level": a more negative curvature $\kappa$ (larger $c$) amplifies the separation of fine-grained motions, whereas smoother curvature helps preserve sentence-level coherence.

Geo-Sign leverages this geometric inductive bias through a novel regularisation framework for a pre-trained mT5 model \cite{xue-etal-2021-mt5}. By projecting skeletal features into hyperbolic space and aligning them with text embeddings via a geometric contrastive loss, we guide the mT5 model to internalize the hierarchical nature of sign language kinematics. Our primary contributions include:
\begin{itemize}[leftmargin=*, itemsep=1pt, topsep=1pt]
    \item \textbf{Hyperbolic Skeletal Representation}: We map multi-part skeletal features, derived from ST-GCNs, into the Poincaré ball using curvature-aware hyperbolic projection layers.
    \item \textbf{Geometric Contrastive Regularisation}: We introduce a contrastive learning objective that operates directly in hyperbolic space, minimizing the geodesic distance between semantically corresponding hyperbolic pose and text embeddings.
    \item \textbf{Hierarchical Aggregation and Alignment Strategies}: We explore two main strategies for this contrastive alignment:
        \begin{enumerate}
            \item A \textit{global semantic alignment} method, which uses a weighted Fréchet mean to aggregate part-specific hyperbolic embeddings into a single global pose representation, then aligns this with a global text embedding.
            \item A \textit{fine-grained part-text alignment} method, which employs a novel hyperbolic attention mechanism. This allows individual pose part embeddings to attend to specific text tokens within the hyperbolic space, generating contextual text embeddings for more detailed contrastive learning.
        \end{enumerate}
\end{itemize}
This geometric regularisation offers several advantages. It aims to inform the mT5 model's understanding by providing representations that inherently respect kinematic hierarchy. The learnable curvature allows the model to adapt the representational space to dataset-specific characteristics. Furthermore, by relying solely on anonymized skeletal data, our approach inherently preserves signer privacy and offers greater computational efficiency compared to methods requiring extensive video processing.

Experiments on the CSL-Daily benchmark \cite{zhou2023gloss} demonstrate Geo-Sign's efficacy. Our skeletal-based approach not only achieves a +1.81 BLEU4 and +3.03\ ROUGE score over state-of-the-art pose-based methods but also matches the performance of comparable vision-based networks. We demonstrate that our method extends to American Sign Language and Isolated Sign Language Recognition. Finally, we also present the first method to surpass SOTA gloss based methods (with respect to the ROUGE score) with a gloss-free approach, highlighting the potential of geometrically-aware representations.

\section{Related Work}
\label{sec:related_work}

Our work intersects with several research areas: Sign Language Translation (SLT), the use of skeletal data for sign and action recognition, and the application of hyperbolic geometry in machine learning.

\subsection{Sign Language Translation (SLT)}
Sign Language Translation aims to bridge the communication gap between Deaf and hearing communities by automatically converting sign language videos into spoken or written language text \cite{Bungeroth2004StatisticalSL,guo-etal-2024-unsupervised, tang2021graph, gong2024signllm, wong2024sign2gpt}. Distinct from Sign Language Recognition (SLR), which often focuses on isolated signs or gloss transcription \cite{camgoz2020sign,  tang2022gloss, zelinka2020neural, Hu_2023_corrnet, rastgoo2021sign}, SLT tackles the more complex task of translating continuous signing across modalities with potentially disparate grammatical structures.

Early SLT methods often involved a two-stage process: recognizing sign glosses (individual lexical units of sign language grammar) and then translating the gloss sequence into the target language \cite{camgoz2020sign, zhou2021improving, hu2021global, hu2022temporal,varol2021read, varol2022scaling, momeni2022automatic, momeni2020watch, momeni2020watch, walsh2024data, walsh2023gloss}. However, this intermediate representation can lead to information loss, while gloss transcriptions are limited in availability. Consequently, end-to-end sequence-to-sequence models have become the dominant paradigm \cite{camgoz2018neural}. Initial approaches utilized Recurrent Neural Networks (RNNs) like LSTMs or GRUs, often with attention mechanisms \cite{stoll2018sign, Guo2018HierarchicalLF}. More recently, Transformer architectures \cite{camgoz2020sign} have demonstrated superior performance in capturing long-range dependencies and context \cite{zuo22_interspeech, sincan2023context, jang2025lost}, enabling direct video-to-text translation \cite{wong2024sign2gpt, guo-etal-2024-unsupervised, duarte2021how2sign, uthus2024ytasl, chen2024c2rl}. Many recent state-of-the-art architectures leverage large pre-trained language models, such as T5 variants, fine-tuned for the task of SLT \cite{chen2022two, zhou2021improving}. These often rely on large pre-trained visual encoders, with incremental improvements seen by upgrading visual backbones from ResNet \cite{Zhou_2023_ICCV}, to I3D \cite{tarres2023sign-instructional}, and more recently to ViT variants like DINO \cite{wong2024sign2gpt, wong2025signrep}. However, as these backbones increase in size, they can limit the number of frames processed concurrently due to quadratically scaling resource demands.

Key challenges in SLT remain, including the scarcity of large-scale annotated datasets \cite{koller2015continuous, camgoz2021content4all, albanie2021bbc}, handling signer variability, modelling linguistic divergence between sign and spoken languages \cite{Wei_2023_ICCV, cheng2023cico}, capturing co-articulation effects \cite{zuo2024improving}, and distinguishing visually similar signs \cite{gan2023contrastive}.

\subsection{Skeletal Representations for Sign Language and Action Recognition}
Using skeletal keypoints, extracted via pose estimation algorithms like OpenPose \cite{cao2019openpose}, MediaPipe \cite{lugaresi2019mediapipe}, or MMPose \cite{, mmpose2020} (in this work we use RTMPose for skeletal features~\cite{Jiang2023RTMPoseRM}), offers several advantages over raw RGB video for sign language analysis. Skeletal data is computationally efficient, robust to background and lighting variations, directly encodes articulation kinematics, enhances privacy by design, and can potentially improve generalization across different signers and environments \cite{zuo2024improving, pu2020boosting, ivashechkin2023improving}.

Graph Convolutional Networks (GCNs) and particularly Spatio-Temporal GCNs (ST-GCNs) have shown great promise by explicitly modelling the spatial structure of the skeleton and its temporal dynamics \cite{yan2018spatial, xiao2020skeleton, duan2022revisiting, xie2018rethinking, qiu2017learning}. However, the quality of skeletal data is heavily dependent on the accuracy of the underlying pose estimation algorithms \cite{10193629}. Furthermore, skeletal data might discard subtle visual cues present in RGB video that could be important for disambiguation. While multi-modal fusion (RGB + pose) has been explored to combine the strengths of both modalities \cite{papadimitriou20_interspeech, tang2021graph, zhou2024scaling, wong2025signrep}, it typically increases computational cost. Our work focuses on enhancing the representational power of skeletal data itself by embedding it in hyperbolic space, aiming to improve its discriminability for SLT without resorting to RGB fusion.

\subsection{Hyperbolic Geometry in Machine Learning}
Hyperbolic geometry, characterized by its constant negative curvature, offers unique properties for representation learning \cite{ganea2018hyperbolic, shimizu2020hyperbolic}. Its most notable feature is the exponential growth of volume with radius, which allows hyperbolic spaces to embed tree-like or hierarchical structures with significantly lower distortion than Euclidean spaces. This makes them particularly suitable for data where such latent hierarchies are believed to exist. Common models of hyperbolic geometry used in machine learning include the Poincaré ball model \cite{nickel2017poincare} and the Lorentz (or hyperboloid) model \cite{nickel2018learning}.

\subsection{Hyperbolic Representation Learning Applications}
The advantageous properties of hyperbolic spaces for modelling hierarchies have led to their successful application in various domains. Hyperbolic Graph Neural Networks (HGNNs) have extended GNN principles to hyperbolic space, demonstrating strong performance on graph-related tasks, especially those involving scale-free or hierarchical graphs \cite{liu2019hyperbolic, yang2022hyperbolic}. In Natural Language Processing (NLP), Poincaré embeddings \cite{nickel2018learning} effectively captured word hierarchies (e.g., WordNet taxonomies), leading to the development of hyperbolic RNNs and Transformers for improved modeling of sequential and relational data \cite{yang2024hypformer}.
Applications in computer vision include hyperbolic Convolutional Neural Networks (CNNs) \cite{bdeir2023fully} and vision-language models that leverage hyperbolic spaces to better align visual and textual concept hierarchies \cite{ibrahimi2024intriguing}.

Our work contributes to this growing body of research by applying hyperbolic representation learning specifically to the domain of skeletal Sign Language Translation. While hyperbolic geometry has been explored for general action recognition from skeletons \cite{franco2023hyperbolic, leng2023dynamic, li2025hyliformer} and in broader NLP contexts \cite{micic2018hyperbolic}, its systematic application to enhance the discriminability of multi-part skeletal features for end-to-end SLT, particularly through a geometric contrastive loss operating in hyperbolic space to regularize a large language model, represents a novel direction. We aim to leverage the geometric properties of the Poincaré ball to refine skeletal representations as they are processed by the language model, thereby improving the translation quality, especially for signs involving fine-grained hierarchical motion.
\section{Methodology}
\label{sec:methodology_balanced_v1}

Geo-Sign regularises a pre-trained mT5 model \cite{xue-etal-2021-mt5} by integrating hyperbolic geometry to capture the hierarchical nature of sign kinematics. We employ the $d_{\text{hyp}}$-dimensional Poincaré ball model, $\mathbb{B}_{c}^{d_{\text{hyp}}} = \{ \mathbf{x} \in \mathbb{R}^{d_{\text{hyp}}} : \|\mathbf{x}\|_2 < 1/\sqrt{c} \}$, with a learnable curvature magnitude $c > 0$. This section first briefly introduces essential hyperbolic operations, then details our pose encoding, hyperbolic projection, and two distinct contrastive alignment strategies.

\subsection{Hyperbolic Geometry Essentials}
\label{subsec:hyperbolic_primer_balanced_v1}
Hyperbolic spaces exhibit exponential volume growth ($V_H(r) \propto e^{(d-1)r}$ for large radius $r$), making them adept at embedding hierarchies with low distortion compared to Euclidean spaces ($V_E(r) \propto r^d$) \cite{nickel2017poincare,ganea2018hyperbolic}. In the Poincaré ball, geometry near the origin ($\|\mathbf{x}\|_2 \approx 0$) is approximately Euclidean, while near the boundary ($\|\mathbf{x}\|_2 \to 1/\sqrt{c}$), distances are magnified, providing capacity to distinguish fine details.

The geodesic distance $d_{\mathbb{B}_c}(\mathbf{u},\mathbf{v})$ between points $\mathbf{u}, \mathbf{v} \in \mathbb{B}_{c}^{d_{\text{hyp}}}$ is:
\begin{equation}
    d_{\mathbb{B}_c}(\mathbf{u},\mathbf{v}) = \frac{2}{\sqrt{c}} \operatorname{artanh}\left( \sqrt{c}\left\| (-\mathbf{u}) \oplus_c \mathbf{v} \right\|_2 \right).
    \label{eq:primer_poincare_dist_balanced_v1}
\end{equation}
This utilizes Möbius addition $\oplus_c$, the hyperbolic analogue of vector addition:
\begin{equation}
    \mathbf{u} \oplus_c \mathbf{v} = \frac{(1+2c\langle\mathbf{u},\mathbf{v}\rangle_2 + c\|\mathbf{v}\|_2^2)\mathbf{u} + (1-c\|\mathbf{u}\|_2^2)\mathbf{v}}{1 + 2c\langle\mathbf{u},\mathbf{v}\rangle_2 + c^2\|\mathbf{u}\|_2^2\|\mathbf{v}\|_2^2}.
    \label{eq:primer_mobius_add_balanced_v1}
\end{equation}
To map Euclidean vectors $\mathbf{v}$ from the tangent space at the origin $\mathcal{T}_{\mathbf{0}}\mathbb{B}_{c}^{d_{\text{hyp}}} \cong \mathbb{R}^{d_{\text{hyp}}}$ into $\mathbb{B}_{c}^{d_{\text{hyp}}}$, we use the exponential map at the origin
$\exp_{\mathbf{0}}^c(\cdot)$:
\begin{equation}
    \exp_{\mathbf{0}}^c(\mathbf{v}) = \tanh\left(\frac{\sqrt{c}\|\mathbf{v}\|_2}{2}\right) \frac{\mathbf{v}}{\frac{\sqrt{c}}{2}\|\mathbf{v}\|_2}, \quad (\mathbf{v} \neq \mathbf{0}).
    \label{eq:primer_exp_map_balanced_v1}
\end{equation}
Its inverse is the logarithmic map at the origin, $\log_{\mathbf{0}}^c(\cdot)$. General maps $\exp_{\mathbf{x}}^c(\cdot)$ and $\log_{\mathbf{x}}^c(\cdot)$ facilitate operations at arbitrary points $\mathbf{x} \in \mathbb{B}_{c}^{d_{\text{hyp}}}$.

\subsection{Skeletal Feature Extraction and Hyperbolic Projection}
\label{ssec:pose_encoding_and_projection_balanced_v1}

\subsubsection{ST-GCN Backbone} We process 2D skeletal keypoints extracted using RTM-Pose \cite{Jiang2023RTMPoseRM}, partitioned into four anatomical groups (body, left/right hands, face). Each group is processed by a part-specific ST-GCN \cite{yan2018spatial} which combines spatial graph convolutions with temporal convolutions to model both joint interdependencies and motion dynamics. Residual connections allow information flow from body joints to hand/face representations.
The ST-GCNs output part-specific feature maps $\mathbf{Z}_p \in \mathbb{R}^{T \times d'_{\text{gcn\_out}}}$ ($T$ is sequence length). For direct input to the mT5 encoder, these are concatenated and linearly projected to $d_{\text{mT5}}$, yielding dynamic Euclidean pose embeddings $\mathbf{E}_{\text{pose}} \in \mathbb{R}^{T \times d_{\text{mT5}}}$. For the hyperbolic regularisation branch, each $\mathbf{Z}_p$ is temporally mean-pooled to a static summary vector $\bar{\mathbf{f}}_p \in \mathbb{R}^{d'_{\text{gcn\_out}}}$, capturing the overall kinematics of part $p$.

\subsubsection{Part-Specific Projection to Poincaré Ball} Each Euclidean summary vector $\bar{\mathbf{f}}_p$ is projected to a hyperbolic embedding $\mathbf{h}_p \in \mathbb{B}_{c}^{d_{\text{hyp}}}$. This projection involves a linear transformation of $\bar{\mathbf{f}}_p$ to dimension $d_{\text{hyp}}$ using a learnable matrix $\mathbf{W}^{p}$, followed by multiplication with a learnable positive scalar $s_p$. This scalar $s_p$ adaptively scales the features in the tangent space, allowing the model to place features from parts with varying motion scales at appropriate "depths" in the hyperbolic space. The resulting tangent vector is then mapped onto the Poincaré ball using the exponential map at the origin (Eq. \ref{eq:primer_exp_map_balanced_v1}):
\begin{equation}
    \mathbf{h}_p = \exp_{\mathbf{0}}^c(s_p \mathbf{W}^{p} \bar{\mathbf{f}}_p).
    \label{eq:method_part_proj_balanced_v1}
\end{equation}
 The set of hyperbolic part embeddings $\{\mathbf{h}_p\}$ forms the input for the subsequent alignment strategies.

\subsection{Hyperbolic Contrastive Loss}
\label{ssec:hyperbolic_alignment_strategies_balanced_v1}
We regularize the mT5 model by minimizing a Geometric Contrastive Loss, adapted from InfoNCE \cite{oord2018representation}, between hyperbolic pose and text embeddings. This loss encourages semantic consistency by pulling corresponding pose-text pairs closer in hyperbolic space while pushing non-corresponding pairs apart. For a batch of $B$ pose embeddings $\{\mathbf{p}_j\}$ and text embeddings $\{\mathbf{t}_j\}$ in $\mathbb{B}_{c}^{d_{\text{hyp}}}$, the loss for a positive pair $(\mathbf{p}_i, \mathbf{t}_i)$ is:
\begin{equation}
\mathcal{L}_{\text{hyp\_pair}}(\mathbf{p}_i, \mathbf{t}_i) = -\log \frac{\exp(-d_{\mathbb{B}_c}(\mathbf{p}_i, \mathbf{t}_i)/\tau)}{\sum_{j=1}^B \exp(-d_{\mathbb{B}_c}(\mathbf{p}_i, \mathbf{t}_j)/\tau + m \cdot \mathbb{I}(i \neq j))}.
\label{eq:method_hyperbolic_contrastive_loss_overall_balanced_v1}
\end{equation}
Here, $\tau > 0$ is a learnable temperature scaling the similarities (negative distances), and $m \ge 0$ is a learnable additive margin for negative pairs. The total regularisation term $\mathcal{L}_{\text{hyp\_reg}}$ is the batch average of $\mathcal{L}_{\text{hyp\_pair}}$.

\subsection{Alignment Methods}

We present two methods for selecting features for alignment which offer benefits and trade-offs. The first takes the geometric mean of the pose and the text embeddings which comes at greater computational efficiency but decreased accuracy. The second method uses poses as individual queries over the text embeddings in hyperbolic space. We then compute the distance between each pose and modified token pair. This improves translation accuracy but incurs additional memory and inference costs (Full details provided in the appendix.)

\subsubsection{Strategy 1: Global Semantic Alignment (Pooled Method)}
This strategy aligns holistic pose and text semantics, promoting high-level understanding.\\
\begin{itemize}[leftmargin=*, itemsep=1pt, topsep=1pt]

    \item\textbf{Pose Embedding ($\mathbf{p}$)}: A global hyperbolic pose $\boldsymbol{\mu}_{\text{pose}} \in \mathbb{B}_{c}^{d_{\text{hyp}}}$ is computed as the weighted Fréchet mean of the part embeddings $\{\mathbf{h}_p\}$. The Fréchet mean is a geometrically sound average in hyperbolic space. Weights $w_p \propto \exp(d_{\mathbb{B}_c}(\mathbf{0}, \mathbf{h}_p))$, normalized via softmax, emphasize parts with more distinct hyperbolic embeddings. The mean is found iteratively (Algorithm \ref{alg:frechet_mean_balanced_v1}) using general logarithmic maps $\log_{\mathbf{x}}^c(\cdot)$ and exponential maps $\exp_{\mathbf{x}}^c(\cdot)$ for tangent space computations.\\
    \item\textbf{Text Embedding ($\mathbf{t}$)}: A global hyperbolic text embedding $\mathbf{h}_{\text{text}} \in \mathbb{B}_{c}^{d_{\text{hyp}}}$ is obtained by mean-pooling Euclidean token embeddings (e.g., from mT5 decoder's final layer) and then projecting this single sentence vector to $\mathbb{B}_{c}^{d_{\text{hyp}}}$ using a hyperbolic projection layer (structurally similar to Eq. \ref{eq:method_part_proj_balanced_v1}).
    \end{itemize}

The contrastive loss $\mathcal{L}_{\text{hyp\_reg}}$ (Eq. \ref{eq:method_hyperbolic_contrastive_loss_overall_balanced_v1}) is then computed between the sets of these global pose embeddings $\{\boldsymbol{\mu}_{\text{pose},i}\}$ and global text embeddings $\{\mathbf{h}_{\text{text},i}\}$.

\begin{algorithm}[htbp]
\caption{Iterative Weighted Fréchet Mean in $\mathbb{B}_{c}^{d_{\text{hyp}}}$}
\label{alg:frechet_mean_balanced_v1}
\begin{algorithmic}[1]
\Require Hyperbolic embeddings $\{\mathbf{h}_p\}_{p=1}^N$, normalized positive weights $\{w_p\}_{p=1}^N$, $c$, $I_{\text{max}}$, $\epsilon_{\text{tol}}$.
\State Initialize $\boldsymbol{\mu}^{(0)} \gets \mathbf{h}_1$ (or other suitable initialization).
\For{$k = 0$ to $I_{\text{max}}-1$}
    \State $\mathbf{v}_{\text{agg}} \gets \mathbf{0} \in \mathcal{T}_{\boldsymbol{\mu}^{(k)}}\mathbb{B}_{c}^{d_{\text{hyp}}}$. \Comment{Aggregated tangent vector at current mean}
    \For{$p=1$ to $N$}
        \State $\mathbf{v}_{\text{agg}} \gets \mathbf{v}_{\text{agg}} + w_p \log_{\boldsymbol{\mu}^{(k)}}^c(\mathbf{h}_p)$. \Comment{Sum weighted log-mapped vectors}
    \EndFor
    \State $\boldsymbol{\mu}^{(k+1)} \gets \exp_{\boldsymbol{\mu}^{(k)}}^c(\mathbf{v}_{\text{agg}})$. \Comment{Update mean via exponential map}
    \State Project $\boldsymbol{\mu}^{(k+1)}$ into $\mathbb{B}_{c}^{d_{\text{hyp}}}$ if numerically necessary.
    \If{$d_{\mathbb{B}_c}(\boldsymbol{\mu}^{(k+1)}, \boldsymbol{\mu}^{(k)}) < \epsilon_{\text{tol}}$} \Comment{Check convergence}
        \State \textbf{break} \Comment{Exit loop on convergence}
    \EndIf
\EndFor
\State $\boldsymbol{\mu}_{\text{pose}} \gets \boldsymbol{\mu}^{(k+1)}$ \Comment{Assign final mean}
\Ensure Estimated Fréchet mean $\boldsymbol{\mu}_{\text{pose}}$.
\end{algorithmic}
\end{algorithm}

\subsubsection{Strategy 2: Fine-Grained Part-Text Alignment (Token Method)}
This strategy aligns individual pose parts (Queries $\bm{h}_p$) with relevant text segments via hyperbolic attention. For each $\bm{h}_p \in \mathbb{B}_{c}^{d_{\text{hyp}}}$ (from Eq.~\ref{eq:method_part_proj_balanced_v1}), a specific context vector $\bm{c}_p \in \mathbb{B}_{c}^{d_{\text{hyp}}}$ is computed.

\begin{itemize}[leftmargin=*, itemsep=1pt, topsep=1pt]
\item\textbf{Tokenization:} First, the sequence of $T$ Euclidean text token embeddings, $\{ \bm{e}_t \}_{t=1}^T$, is projected into the Poincaré ball. This yields a sequence of hyperbolic embeddings $V = \{ \bm{v}_t \}_{t=1}^T$, which serve as the values and hold the original semantic meaning of each token.

\item\textbf{Key Transformation:} The value sequence $V$ is transformed by a learnable Möbius affine transformation (using $\bm{M}$ and $\bm{b}$) to create a sequence of keys $K = \{ \bm{k}_t \}_{t=1}^T$:

\begin{equation}
\label{eq:key_transform}
\bm{k}_t = (\bm{M} \otimes \bm{v}_t) \oplus \bm{b}
\end{equation}

\item\textbf{Attention Scores:} Scores $s_{pt}$ are computed as the negative geodesic distance between each pose query $\bm{h}_p$ and each text key $\bm{k}_t$. A smaller distance signifies greater relevance.
\begin{equation}
\label{eq:attn_scores}
s_{pt} = -d_{\mathbb{B}_c}(\bm{h}_p, \bm{k}_t)
\end{equation}

\item\textbf{Context Vector:} These scores are normalized via softmax (with masking for padding) to produce attention weights $\alpha_{pt}$. The final context vector $\bm{c}_p$ is the hyperbolic weighted midpoint ($\mu$) of the original values $V$ weighted by the pose-shifted attention embeddings:
\begin{equation}
\label{eq:context_vector}
\bm{c}_p = \mu_{\mathcal{B}_c}\left(\{ \bm{v}_t \}_{t=1}^T, \{ \alpha_{pt} \}_{t=1}^T\right)
\end{equation}
The final $\mathcal{L}_{\text{hyp\_reg}}$ is the average of $K$ individual contrastive losses (Eq.~\ref{eq:method_hyperbolic_contrastive_loss_overall_balanced_v1}), one for each $(\bm{h}_p, \bm{c}_p)$ alignment pair.
\end{itemize}
\subsection{Training Objective and Optimization}
\label{ssec:training_objective_balanced_v1}
The model is trained end-to-end by minimizing the total loss $\mathcal{L}_{\text{total}} = \alpha \cdot \mathcal{L}_{\text{CE}} + (1-\alpha) \cdot \mathcal{L}_{\text{hyp\_reg}}$. This combines the standard cross-entropy translation loss $\mathcal{L}_{\text{CE}}$ (with label smoothing) with the hyperbolic regularisation term $\mathcal{L}_{\text{hyp\_reg}}$ from one of the alignment strategies. The blending factor $\alpha \in [0.1, 1.0]$ is dynamically adjusted during training via a learnable parameter and training progress, allowing an initial focus on $\mathcal{L}_{\text{hyp\_reg}}$ before increasing the influence of $\mathcal{L}_{\text{CE}}$.

Optimization employs AdamW \cite{kingma2014adam, loshchilov2017decoupled} for Euclidean parameters (ST-GCNs, mT5, linear layers), with learning rate  $3 \times 10^{-5}$. Hyperbolic parameters, including the learnable curvature $c$ (optimized in log-space, e.g., $\log c$) and manifold-constrained parameters, use Riemannian Adam (RAdam) \cite{becigneul2023riemannian} with a comparable learning rate. RAdam adapts updates to the manifold's geometry by operating in tangent spaces. All hyperbolic computations utilize high-precision floating-point numbers (e.g., `float32`) for numerical stability. A key stabilization step before applying any exponential map $\exp_{\mathbf{x}}^c(\mathbf{v})$ involves projecting the input tangent vector $\mathbf{v}$ via $\mathbf{v} \leftarrow \mathbf{v} / \max(1, \sqrt{c}\|\mathbf{v}\|_2 + \epsilon)$ for a small $\epsilon > 0$ (e.g., $10^{-5}$), ensuring the argument is well-behaved and the output point remains strictly within the Poincaré ball.

\section{Experiments}
\label{sec:experiments_final}

We evaluate Geo-Sign on Chinese Sign Language (CSL) and American Sign Language (ASL). For CSL we use the CSL-Daily dataset \cite{zhou2021improving, zhou2023gloss}, a large-scale corpus for Chinese Sign Language to Chinese text translation, comprising over 20,000 videos. For American Sign Language we perform translation experiments on How2Sign \cite{duarte2021how2sign} and isolated sign language recognition experiments on WLASL2000 \cite{li2020wlasl}.

Translation quality is assessed using BLEU \cite{papineni2002bleu} (B-1, B-4) and ROUGE-L \cite{lin2004rouge} (R-L) scores where a higher percentage represents a more accurate translation.

\subsection{Experimental Setup} 
Our framework builds upon the Uni-Sign architecture \cite{li2025uni}, using its pre-trained ST-GCN weights (trained on skeletal features from the CSL-News dataset \cite{li2025uni} for CSL and the YTASL \cite{uthus2024ytasl} dataset for ASL) and an mT5 model \cite{xue-etal-2021-mt5} as the language decoder. Following Uni-Sign's fine-tuning protocol, which involves 40 epochs of supervised finetuning on CSL-Daily or YTASL, with fused skeletal and RGB features, we remove the RGB encoder and instead apply our hyperbolic regularisation. This allows for a fair comparison of the impact of our geometric regularisation. We investigate both the "Pooled Method" (Strategy 1) and the "Token Method" (Strategy 2) for hyperbolic alignment. To assess the specific contribution of hyperbolic geometry, we also compare against a "Euclidean regularisation" baseline, where the contrastive loss operates on Euclidean projections to the Poincaré ball where curvature is minimal (0.001) and approximately Euclidean. Key hyperparameters for the hyperbolic components (initial curvature $c=1.5$, dimension $d_{\text{hyp}}=256$, and $\alpha=0.70$) are minimally tuned on the development set (further details in the appendix).

\subsection{Results on Chinese Sign Language (CSL)}
\label{ssec:quantitative_results_final}
\Cref{tab:csl_daily_results_final} presents our main results on the CSL-Daily test set, comparing Geo-Sign with prior art and baselines. Our Geo-Sign (Hyperbolic Token) model, using only pose data, achieves a test BLEU-4 of 27.42\% and ROUGE-L of 57.95\%. This represents a significant improvement of +1.81 BLEU-4 and +3.03 ROUGE-L over the strong Uni-Sign (Pose) baseline (25.61\% BLEU-4, 54.92\% ROUGE-L). Notably, this performance surpasses all other reported gloss-free pose-only methods and is competitive with, or exceeds, several RGB-only and even some gloss-based methods, underscoring the efficacy of our geometric regularisation. The Geo-Sign (Hyperbolic Pooled) variant also outperforms the Euclidean regularisation methods and the Uni-Sign pose baseline, demonstrating the general benefit of hyperbolic geometry. The "Euclidean Token" regularisation already shows improvement over the Uni-Sign baseline, suggesting the contrastive alignment itself is beneficial, but the further gains from hyperbolic geometry are substantial.

\subsubsection{Results on American Sign Language}

In Table~\ref{tab:how2sign_slt} we show results on Sign Language Translation for American Sign Language on the How2Sign \cite{duarte2021how2sign} dataset. Our method shows increased performance over all pose based methods but performs marginally worse than the best RGB method \cite{rust-etal-2024-ssvp} which benefits from a longer pre-training duration and scale. In Table~\ref{tab:wlasl_slr} we also compare our approach on Isolated Sign Language Recognition (ISLR) with the WLASL2000 \cite{li2020wlasl} dataset. For isolated recognition our method shows a small improvement in Top-1 Accuracy for both instance (+0.12) and class-level (+0.57).

\begin{table*}[t]
\centering
\label{tab:csl_daily_results_final}
\resizebox{\textwidth}{!}{%
\begin{tabular}{l|cc|ccc|ccc}
\toprule
\multirow{2}{*}{\textbf{Method}} & \multicolumn{2}{c|}{\textbf{Modality}} & \multicolumn{3}{c|}{\textbf{Dev Set}} & \multicolumn{3}{c}{\textbf{Test Set}} \\
\cmidrule(lr){2-3}\cmidrule(lr){4-6}\cmidrule(lr){7-9}
& Pose & RGB & B-1 & B-4 & R-L & B-1 & B-4 & R-L \\

\midrule
\rowcolor[gray]{.9}\multicolumn{9}{c}{\textit{Gloss-Based Methods (Prior Art)}} \\
SLRT~\cite{camgoz2020sign} & -- & \checkmark & 37.47 & 11.88 & 37.96 & 37.38 & 11.79 & 36.74 \\
TS-SLT~\cite{chen2022two} & \checkmark & \checkmark & 55.21 & 25.76 & 55.10 & 55.44 & 25.79 & 55.72 \\
CV-SLT~\cite{zhao2024conditional} & -- & \checkmark & -- & \underline{28.24} & 56.36 & \underline{58.29} & \underline{28.94} & 57.06 \\
\midrule

\rowcolor[gray]{.9}\multicolumn{9}{c}{\textit{Gloss-Free Methods (Prior Art)}} \\
MSLU~\cite{zhou2024scaling} & \checkmark & -- & 33.28 & 10.27 & 33.13 & 33.97 & 11.42 & 33.80 \\
SLRT~\cite{camgoz2020sign} (Gloss-Free variant) & -- & \checkmark & 21.03 &  4.04 & 20.51 & 20.00 &  3.03 & 19.67 \\
GASLT~\cite{yin2023gloss} & -- & \checkmark & -- & -- & -- & 19.90 &  4.07 & 20.35 \\
GFSLT-VLP~\cite{zhou2023gloss} & -- & \checkmark & 39.20 & 11.07 & 36.70 & 39.37 & 11.00 & 36.44 \\
FLa-LLM~\cite{chen2024factorized} & -- & \checkmark & -- & -- & -- & 37.13 & 14.20 & 37.25 \\
Sign2GPT~\cite{wong2024sign2gpt} & -- & \checkmark & -- & -- & -- & 41.75 & 15.40 & 42.36 \\
SignLLM~\cite{gong2024signllm} & -- & \checkmark & 42.45 & 12.23 & \textbf 39.18 & 39.55 & 15.75 & 39.91 \\
C$^{2}$RL~\cite{chen2024c2rl} & -- & \checkmark & -- & -- & -- & 49.32 & 21.61 & 48.21 \\
\midrule
\rowcolor[gray]{.9}\multicolumn{9}{c}{\textit{Our Models and Baselines}} \\
Uni-Sign \cite{li2025uni} (Pose) & \checkmark & -- & 53.24 & 25.27 & 54.34 & 53.86 & 25.61 & 54.92 \\
Uni-Sign \cite{li2025uni} (Pose+RGB) & \checkmark & \checkmark & 55.30 & 26.25 & 56.03 & 55.08 & 26.36 & 56.51 \\
\midrule
\textbf{Geo-Sign (Euclidean Pooled)} & \checkmark & -- & 53.53 & 25.78 & 55.38 & 53.06 & 25.72 & 55.57 \\
\textbf{Geo-Sign (Euclidean Token)} & \checkmark & -- & 53.93 & 25.91 & 55.20 & 54.02 & 25.98 & 53.93 \\
\midrule
\textbf{Geo-Sign (Hyperbolic Pooled)} & \checkmark & -- & 55.19 & 26.90 & 56.93 & 55.80 & 27.17 & 57.75 \\
\textbf{Geo-Sign (Hyperbolic Token)} & \checkmark & -- & \textbf{55.57} & \textbf{27.05} & \textbf{57.27} & \textbf{55.89} & \textbf{27.42} & \textbf{57.95} \\
\bottomrule
\end{tabular}
}
\caption{Sign Language Translation performance on the CSL-Daily dataset. BLEU scores (B-1, B-4) and ROUGE-L (R-L) are reported as percentages (\%). Higher is better. `Pose' and `RGB' indicate input modalities. Uni-Sign is the base architecture sharing pre-training/fine-tuning setups but without our regularisation. Euclidean regularisation applies contrastive loss in Euclidean space. Our Hyperbolic Token method surpasses all other pose-only methods and is competitive with top RGB/multimodal methods. Gloss-based methods that outperform our method are underlined.}
\end{table*}

\begin{table*}[t]
\centering
\small 

\begin{minipage}[t]{0.51\textwidth} 
    \centering
    \setlength{\tabcolsep}{4pt}
        \captionof{table}{Sign Language Translation (SLT) results on the How2Sign dataset. Metrics are BLEU (B-1, B-4) and ROUGE-L (R-L). Higher is better.}
    \begin{tabular}{l|cc|ccc}
    \toprule
    \multirow{2}{*}{\textbf{Method}} & \multicolumn{2}{c|}{\textbf{Modality}} & \multicolumn{3}{c}{\textbf{Test Set}} \\
    \cmidrule(lr){2-3} \cmidrule(lr){4-6}
    & Pose & RGB & B-1 & B-4 & R-L \\
    \midrule
    \rowcolor[gray]{.9} \multicolumn{6}{c}{\textit{Gloss-Free Methods (Prior Art)}} \\
    GloFE-VN~\cite{lin-etal-2023-GLoFE} & \checkmark & -- & 14.9 & 2.2 & 12.6 \\
    YouTube-ASL~\cite{uthus2024ytasl} & \checkmark & -- & 37.8 & 12.4 & - \\
    MSLU~\cite{zhou2024scaling} & \checkmark & -- & 20.1 & 2.4 & 17.2 \\
    SLT-IV~\cite{tarres2023sign-instructional} & -- & \checkmark & 34.0 & 8.0 & - \\
    C$^2$RL~\cite{chen2024c2rl} & -- & \checkmark & 29.1 & 9.4 & 27.0 \\
    FLa-LLM~\cite{chen2024factorized} & -- & \checkmark & 29.8 & 9.7 & 27.8 \\
    SignMusketeers~\cite{gueuwou-etal-2025-signmusketeers} & -- & \checkmark & \underline{41.5} & 14.3 & - \\
    SSVP-SLT~\cite{rust-etal-2024-ssvp} & -- & \checkmark & \textbf{43.2} & \textbf{15.5} & \textbf{38.4} \\
    \midrule
    \rowcolor[gray]{.9} \multicolumn{6}{c}{\textit{Our Models and Baselines}} \\
    Uni-Sign (Pose) & \checkmark & -- & 40.4 & 14.5 & 34.3 \\
    Uni-Sign (Pose+RGB) & \checkmark & \checkmark & 40.2 & 14.9 & \underline{36.0} \\
    \textbf{Geo-Sign (Token)} & \checkmark & -- & 40.8 & \underline{15.1} & 35.4 \\
    \bottomrule
    \end{tabular}
    \label{tab:how2sign_slt}
\end{minipage}
\hfill 
\begin{minipage}[t]{0.46\textwidth} 
    \centering
    \setlength{\tabcolsep}{4pt}
     \captionof{table}{Isolated Sign Language Recognition (ISLR) results on WLASL2000. We report Top-1 Accuracy for Per-Instance (P-I) and Per-Class (P-C). $^\dagger$ from \cite{hu2021signbert}.}
    \begin{tabular}{l|cc|cc}
    \toprule
    \multirow{2}{*}{\textbf{Method}} & \multicolumn{2}{c|}{\textbf{Mod.}} & \multicolumn{2}{c}{\textbf{Test Acc. (\%)}} \\
    \cmidrule(lr){2-3} \cmidrule(lr){4-5}
    & P & RGB & P-I & P-C \\
    \midrule
    \rowcolor[gray]{.9} \multicolumn{5}{c}{\textit{Prior Art}} \\
    ST-GCN$^{\dagger}$~\cite{yan2018stgcn} & \checkmark & -- & 34.40 & 32.53 \\
    SignBERT~\cite{hu2021signbert} & \checkmark & -- & 39.40 & 36.74 \\
    HMA~\cite{hu2021hand} & -- & \checkmark & 37.91 & 35.90 \\
    BEST~\cite{zhao2023best} & \checkmark & -- & 46.25 & 43.52 \\
    SignBERT+~\cite{hu2023signbert+} & \checkmark & -- & 48.85 & 46.37 \\
    MSLU~\cite{zhou2024scaling} & \checkmark & -- & 56.29 & 53.29 \\
    NLA-SLR~\cite{zuo2023natural} & \checkmark & \checkmark & 61.05 & 58.05 \\
     Sign-Rep~\cite{wong2025signrep} & \checkmark & \checkmark & 61.05 & 58.89 \\
    \midrule
    \rowcolor[gray]{.9} \multicolumn{5}{c}{\textit{Our Models}} \\
    Uni-Sign (Pose) & \checkmark & -- & 63.13 & 60.90 \\
    Uni-Sign (Pose+RGB) & \checkmark & \checkmark & \underline{63.52} & \underline{61.32} \\
    \textbf{Geo-Sign (Token)} & \checkmark & --- & \textbf{63.64} & \textbf{61.89} \\
    \bottomrule
    \end{tabular}
    \label{tab:wlasl_slr}
\end{minipage}
\end{table*}

\subsubsection{Ablation Studies}
Ablation studies on the CSL-Daily test set for our best performing Geo-Sign (Hyperbolic Token) model are presented in \Cref{tab:ablations_combined_horizontal}. We investigate the impact of the initial hyperbolic curvature $c$ and the loss blending factor $\alpha$.
For curvature $c$ (with $\alpha=0.7$), setting $c=0.001$ effectively makes the projection Euclidean (as $\tanh(x) \approx x$ for small $x$, which means almost zero hyperbolic warping). We observe that increasing curvature from this Euclidean-like baseline ($c=0.001$, BLEU-4 25.91\%) generally improves performance, with optimal BLEU-4 (27.42\%) achieved at $c=1.5$. ROUGE-L peaks at $c=2.0$ (58.08\%), though BLEU-4 slightly dips to 27.25\%, suggesting a trade-off. This indicates that a significant degree of negative curvature is beneficial for capturing sign language structure.
For the loss blending factor $\alpha$ (with $c=1.5$), a value of $\alpha=0.7$ (i.e., 30\% weight to the hyperbolic loss) yields the best BLEU-4 (27.42\%) and ROUGE-L (57.95\%). Lower or higher $\alpha$ values result in decreased performance, indicating that the hyperbolic regularisation provides a substantial complementary signal to the primary translation loss, but should not entirely dominate it during the 40 epochs of fine-tuning. Finally, we assess the robustness of our approach to pose perturbation and poor pose estimation. To do so, we add Gaussian noise to the pose embeddings before embedding via the ST-GCN. Our method shows how hyperbolic regularisation improves robustness to pose noise compared to the Uni-Sign Euclidean baseline. We attribute this to the larger geodesic margins between pose embeddings in the hyperbolic space, making the model less susceptible to noisy perturbation.

\begin{table*}[t]
\centering
\footnotesize 
\caption{Ablation studies for Geo-Sign on the CSL-Daily test set, examining (a) initial curvature $c$, (b) loss blending factor $\alpha$, and (c) robustness of poses to Gaussian noise. We demonstrate that hyperbolic regularisation improves robustness to pertubation and poor pose estimation.}
\label{tab:ablations_combined_horizontal}

\subcaptionbox{Impact of Curvature $c$. \label{subtab:curvature_ablation}}{%
    \begin{tabular}{ccc}
    \toprule
    Curvature ($c$) & B-4 & R-L \\
    \midrule
    0.00 (Euclidean) & 25.98 & 53.93 \\
    0.10 & 26.56 & 57.56 \\
    0.50 & 26.34 & 56.30 \\
    1.00 & 27.04 & 57.67 \\
    \textbf{1.50} & \textbf{27.42} & 57.95 \\
    2.00 & 27.25 & \textbf{58.08} \\
    \bottomrule
    \end{tabular}
}
\hfill 
\subcaptionbox{Impact of $\alpha$. \label{subtab:alpha_ablation}}{%
    \begin{tabular}{ccc}
    \toprule
    $\alpha$ & B-4 & R-L \\
    \midrule
    0.10 & 25.74 & 56.20 \\
    0.50 & 26.79 & 57.38 \\
    \textbf{0.70} & \textbf{27.42} & \textbf{57.95} \\
    0.90 & 26.92 & 57.67 \\
    \bottomrule
    \end{tabular}
}
\hfill 
\subcaptionbox{Noise Robustness (B-4). \label{subtab:noise_robustness}}{%
    \begin{tabular}{lcc}
    \toprule
    Noise & Geo-Sign & Uni-Sign \\
    \midrule
    0.00 & 27.42 & 26.25 \\
    0.01 & 26.30 \color{gray}{(-4\%)}  & 24.14 \color{gray}{(-8\%)} \\
    0.02 & 24.60 \color{gray}{(-10\%)} & 21.50 \color{gray}{(-18\%)} \\
    0.03 & 19.07 \color{gray}{(-30\%)} & 14.40 \color{gray}{(-45\%)} \\
    0.04 & 11.63 \color{gray}{(-58\%)} & 7.20  \color{gray}{(-73\%)} \\
    0.05 & 5.98  \color{gray}{(-78\%)} & 3.01  \color{gray}{(-89\%)} \\
    \bottomrule
    \end{tabular}
}
\end{table*}

\subsection{Qualitative Analysis: Visualizing Embedding Spaces}
\label{ssec:qualitative_analysis_final}
\begin{figure}[htpb]
    \centering
    \includegraphics[width=1\linewidth]{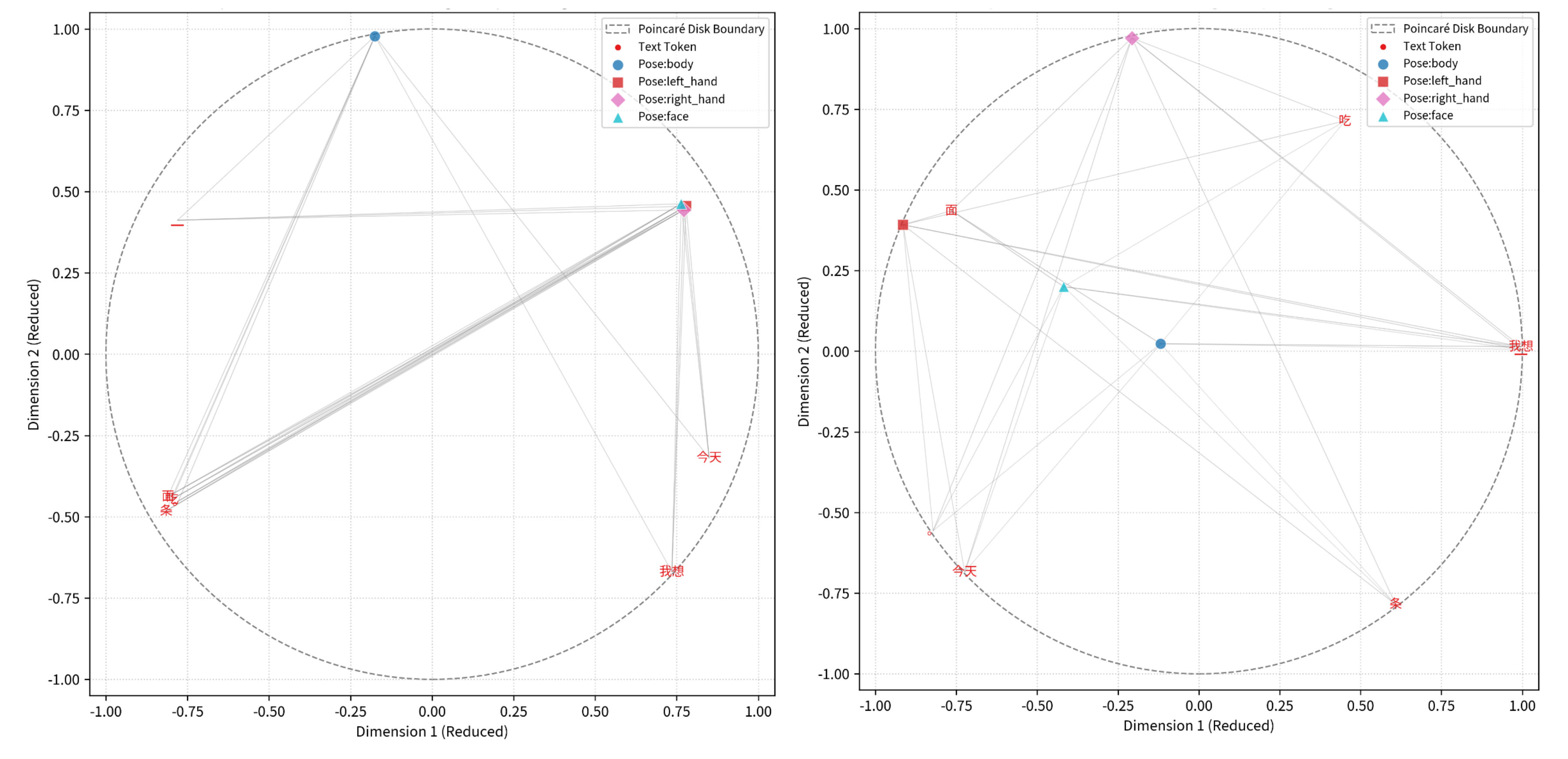} 
    \caption{UMAP projection of pose part summary embeddings ($\bar{\mathbf{f}}_p$ onto the 2D Poincaré disk). (\textbf{Left}) Embeddings from the Euclidean Token regularisation model ($c=0.001$). (\textbf{Right}) Embeddings from the Geo-Sign (Hyperbolic Token) model. The hyperbolic embeddings show a more structured distribution, with hand features (representing finer details) often pushed towards the periphery indicative of a learned kinematic hierarchy.}
    \label{fig:hyperbolic_proj_final}
\end{figure}

To intuitively understand the effect of hyperbolic regularisation, we visualise the learned pose embeddings. \Cref{fig:hyperbolic_proj_final} shows UMAP \cite{mcinnes2018umap} projections of these embeddings into the 2D Poincaré disk (by log-mapping hyperbolic embeddings to the tangent space at the origin, then applying UMAP). We compare embeddings from our Geo-Sign (Hyperbolic Token) model against those from the Geo-Sign (Euclidean Token) model, which uses the same contrastive token-level alignment but without hyperbolic projection (curvature $c=0$).

The Euclidean embeddings (\Cref{fig:hyperbolic_proj_final}, Left) appear relatively clustered and undifferentiated. In contrast, the hyperbolic embeddings (\Cref{fig:hyperbolic_proj_final}, Right) exhibit a more structured distribution. Notably, embeddings corresponding to hand articulations (often carrying fine-grained lexical information) tend to occupy regions further from the origin, towards the periphery of the Poincaré disk. This is consistent with hyperbolic geometry's property of expanding space near the boundary, providing more capacity to distinguish subtle variations. Conversely, features representing larger body movements or overall posture (often conveying prosodic or grammatical information) tend to be located more centrally. This visualised structure suggests that the hyperbolic model indeed learns to place features in a manner that reflects the hierarchical nature of sign kinematics, with fine details pushed to high-curvature regions and global features remaining near the low-curvature origin.

\section{Conclusion}
\label{sec:conclusion_final}

This paper introduced Geo-Sign, a novel framework that enhances Sign Language Translation by leveraging hyperbolic geometry to model the inherent hierarchical structure of sign language kinematics. By projecting skeletal features from ST-GCNs into the Poincaré ball and employing a geometric contrastive loss, Geo-Sign regularises a pre-trained mT5 model, guiding it to learn more discriminative and geometrically aware representations. We explored two alignment strategies: a global pooled method and a fine-grained token-based attention method operating directly in hyperbolic space. Our experimental results on both Chinese Sign Language and American Sign Language demonstrate the significant benefits of this approach.

\section*{Acknowledgements}
This work was supported by the SNSF project ‘SMILE II’ (CRSII5 193686), the Innosuisse IICT Flagship (PFFS-21-47), EPSRC grant APP24554 (SignGPT-EP/Z535370/1) and through funding from Google.org via the AI for Global Goals scheme. This work reflects only the author’s views and the funders are not responsible for any use that may be made of the information it contains.

Thank you to Low Jian He for reviewing the Chinese text translations.

\clearpage 
\bibliographystyle{plainnat}
\bibliography{neurips_2025}           

\appendix
\clearpage
\hrule 
\begin{center}
    \Large\textbf{Appendix}
\end{center}
\vspace{0.5em}

\startcontents[app]              
\printcontents[app]{}{1}{\setcounter{tocdepth}{2}}
\clearpage 
\vspace{1em}
\hrule
\section{Introduction}
\label{sec:sup_intro}
In this appendix, we provide comprehensive supplementary details to accompany our main paper. The goal is to offer an in-depth understanding of our methodology, experimental setup, and the underlying geometric principles, thereby ensuring clarity and facilitating the reproducibility of our work.

This document elaborates on:
\begin{itemize}[noitemsep,topsep=3pt,parsep=3pt,partopsep=0pt]
    \item The specifics of pose feature extraction and the Spatio-Temporal Graph Convolutional Network (ST-GCN) architecture employed (\Cref{app:pose_stgcn}).
    \item Detailed explanations and implementations of our proposed hyperbolic alignment strategies, including the Pooled Method and the Token Method (\Cref{app:hyperbolic_alignment_strategies_detail}).
    \item Further mathematical derivations and discussions pertinent to hyperbolic operations, such as Fréchet mean computation and contrastive loss gradients (\Cref{sec:supp_math}).
    \item Elaboration on the learnable parameters within our model, particularly the manifold curvature $c$ and the loss blending factor $\alpha$ (\Cref{app:learnable_model_params}).
    \item A discussion of computational considerations, experimental setup, and qualitative results (\Cref{app:exp_setup_analysis}).
    \item Key code snippets for essential components of Geo-Sign are provided in \Cref{sec:code_listings} to aid in understanding and replication.
\end{itemize}

\section{Hyperbolic Geometry Preliminaries: A Brief Refresher}
\label{app:hyperbolic_preliminaries}
To ensure this supplementary material is self-contained and accessible, this section briefly recaps key concepts from hyperbolic geometry, as introduced in Section~3.1 (``Hyperbolic Geometry Essentials'') of the main paper. 

We operate within the $d_{\text{hyp}}$-dimensional Poincaré ball model, denoted $\mathbb{B}_{c}^{d_{\text{hyp}}} = \{ \mathbf{x} \in \mathbb{R}^{d_{\text{hyp}}} : \|\mathbf{x}\|_2 < 1/\sqrt{c} \}$. This space is characterised by a constant negative curvature $\kappa = -c$, where $c > 0$ is a learnable parameter representing the magnitude of the curvature.

The Poincaré ball model is chosen for its conformal nature, where angles are preserved locally, and its intuitive representation of hyperbolic space within a Euclidean unit ball (scaled by $1/\sqrt{c}$). Key operations include:

\begin{itemize}[noitemsep,topsep=3pt,parsep=3pt,partopsep=0pt]
    \item \textbf{Geodesic Distance} $d_{\mathbb{B}_c}(\mathbf{u},\mathbf{v})$: This is the shortest path between two points $\mathbf{u}, \mathbf{v}$ within the curved space of the Poincaré ball. It is formally defined in Eq.~(1) of the main paper. Unlike Euclidean distance, it expands significantly as points approach the boundary of the ball.
    \item \textbf{Möbius Addition} $\mathbf{u} \oplus_c \mathbf{v}$: This operation is the hyperbolic analogue of vector addition in Euclidean space, defined in Eq.~(2) of the main paper (consistent with formulations in, e.g., \cite{ganea2018hyperbolic}). It is essential for defining translations and other transformations in hyperbolic space while respecting its geometry.
    \item \textbf{Exponential Map} $\exp_{\mathbf{x}}^c(\mathbf{v})$: This map takes a tangent vector $\mathbf{v}$ residing in the tangent space $\mathcal{T}_{\mathbf{x}}\mathbb{B}_{c}^{d_{\text{hyp}}}$ at a point $\mathbf{x}$ on the manifold and maps it to another point on the manifold along a geodesic. The map from the origin, $\exp_{\mathbf{0}}^c(\cdot)$ (Eq.~(3), main paper), is particularly important as it projects Euclidean feature vectors (which can be considered as residing in $\mathcal{T}_{\mathbf{0}}\mathbb{B}_{c}^{d_{\text{hyp}}}$) into the Poincaré ball.
    \item \textbf{Logarithmic Map} $\log_{\mathbf{x}}^c(\mathbf{y})$: This is the inverse of the exponential map. It takes two points $\mathbf{x}, \mathbf{y}$ on the manifold and returns the tangent vector at $\mathbf{x}$ that points along the geodesic towards $\mathbf{y}$.
    \item \textbf{Möbius Transformations}: These are isometries (distance-preserving transformations) of hyperbolic space. In our work, we use learnable Möbius transformations, such as Möbius matrix-vector products ($\mathbf{M} \otimes_c \mathbf{v} = \exp_{\mathbf{0}}^c(\mathbf{M} \log_{\mathbf{0}}^c(\mathbf{v}))$) and Möbius bias additions, to implement affine-like transformations within our hyperbolic attention mechanism.
\end{itemize}
These tools allow us to define neural network operations directly within hyperbolic space. As with all hyperbolic operations in the paper, we utilise the geoopt library \cite{geoopt2020kochurov} in Pytorch.

\section{Methodology Details}
\label{app:methodology_details}

\subsection[Pose Extraction and ST-GCN Architecture]{Pose Extraction and ST-GCN Architecture Details}
\label{app:pose_stgcn}
Our Geo-Sign framework utilizes skeletal pose data as input. This section details the extraction process and the architecture of the Spatio-Temporal Graph Convolutional Networks (ST-GCNs) used to encode this data.

\subsubsection{Pose Data Source and Preprocessing}
We use the 2D skeletal keypoints provided by the UniSign \cite{li2025uni} framework, which were originally extracted using RTMPose-X \cite{Jiang2023RTMPoseRM} based on the COCO-WholeBody keypoint definition \cite{jin2020whole}. The keypoints are organised into four distinct anatomical groups for targeted processing:
\begin{itemize}[noitemsep,topsep=3pt,parsep=3pt,partopsep=0pt]
    \item \textbf{Body}: Includes 9 joints (COCO indices 1, 4--11).
    \item \textbf{Left Hand}: Includes 21 joints (COCO indices 92--112).
    \item \textbf{Right Hand}: Includes 21 joints (COCO indices 113--133).
    \item \textbf{Face}: Includes 16 keypoints from the facial region (COCO indices 24, 26, 28, 30, 32, 34, 36, 38, 40, 54, 84--91).
\end{itemize}
For normalization, specific anchor joints are used for hand and face parts: joint 92 (left wrist) for the left hand, joint 113 (right wrist) for the right hand, and joint 54 (a central face point) for the face. The body part features are not anchor-normalised in this scheme to preserve global torso positioning.

\subsubsection{ST-GCN Architecture}
Each anatomical group is processed by a dedicated ST-GCN stream, following the methodology of Yan et al. \cite{yan2018spatial}. The ST-GCN is adept at learning representations from skeletal data by explicitly modeling spatial joint relationships and temporal motion dynamics.

The core of the ST-GCN involves:
\begin{enumerate}[noitemsep,topsep=3pt,parsep=3pt,partopsep=0pt]
    \item \textbf{Graph Definition}: The skeletal structure for each part is defined as a graph, where joints are nodes and natural bone connections are edges. The \texttt{Graph} class, detailed in \Cref{lst:gcn_utils_graph} (\Cref{sec:code_listings}), handles the construction of these graphs and their corresponding adjacency matrices.
    \item \textbf{Initial Projection}: Input keypoint coordinates are first linearly projected to a higher-dimensional feature space using a linear layer (referred to as \texttt{proj\_linear} in our codebase).
    \item \textbf{ST-GCN Blocks}: A sequence of ST-GCN blocks processes these features. Each block (see \texttt{STGCN\_block} in \Cref{lst:stgcn_block_chain}, \Cref{sec:code_listings}) consists of:
        \begin{itemize}[noitemsep,topsep=3pt,parsep=3pt,partopsep=0pt]
            \item A \textbf{Spatial Graph Convolution (SGC)} layer, which aggregates information from neighboring joints. The operation for a node (joint) $v_i$ at layer $(l)$ can be expressed generally as:
            \begin{equation} \label{eq:supp_sgc_ref_v3_corrected_final}
                \mathbf{f}_{\text{out}}(v_i)^{(l)} = \sum_{k=1}^{K} \left( \sigma \left( \mathbf{A}_k \mathbf{X}^{(l)} \mathbf{W}_k^{(l)} \right) \right)_{i},
            \end{equation}
            where $\mathbf{X}^{(l)} \in \mathbb{R}^{N \times C_{in}}$ is the matrix of input features for $N$ nodes with $C_{in}$ channels, $\mathbf{W}_k^{(l)} \in \mathbb{R}^{C_{in} \times C_{out}}$ are learnable weight matrices for the $k$-th kernel transforming node features to $C_{out}$ channels. $\mathbf{A}_k \in \mathbb{R}^{N \times N}$ is the adjacency matrix for the $k$-th spatial kernel, defining the neighborhood aggregation based on chosen strategies (we use the spatial configuration partitioning as in the original ST-GCN paper \cite{yan2018spatial}). $\sigma$ is an activation function (ReLU in our case), and $(\cdot)_i$ denotes selection of the $i$-th row (features for node $v_i$). The precise implementation involving tensor reshaping and \texttt{einsum} for efficient aggregation over multiple adjacency kernels is detailed in the \texttt{GCN\_unit} code in \Cref{lst:stgcn_block_chain}.
            
            \item A \textbf{Temporal Convolutional Network (TCN)} layer, which applies 1D convolutions across the time dimension to model motion patterns.
        \end{itemize}
    \item \textbf{Residual Connections}: To allow richer feature interaction, residual connections are introduced from the body stream's ST-GCN output to the hand and face streams before their final temporal fusion layers. This allows global body posture context to inform the interpretation of fine-grained hand and face movements. Details are in \Cref{lst:models_residual_gcn} (\Cref{sec:code_listings}). This design choice treats body features as fixed contextual input for the parts during each forward pass, isolating the body feature extractor from direct updates via part-specific losses.
\end{enumerate}
The output of each part-specific ST-GCN stream is a feature map $\mathbf{Z}_p \in \mathbb{R}^{T \times d'_{\text{gcn\_out}}}$, where $T$ is the sequence length and $d'_{\text{gcn\_out}}$ is the GCN output feature dimension. For the hyperbolic regularization branch, these $\mathbf{Z}_p$ are temporally mean-pooled to produce static summary vectors $\bar{\mathbf{f}}_p \in \mathbb{R}^{d'_{\text{gcn\_out}}}$, which encapsulate the overall kinematics of part $p$ for subsequent hyperbolic projection. 

\subsection[Hyperbolic Alignment Strategies]{Hyperbolic Alignment Strategies: Detailed Implementation}
\label{app:hyperbolic_alignment_strategies_detail}
This section provides a more detailed explanation of the two hyperbolic alignment strategies introduced in Section~3.3 of the main paper. These strategies are designed to regularize the mT5 model by aligning pose and text representations within the Poincaré ball.

\subsubsection{Pooled Method (Global Semantic Alignment)}
\label{app:pooled_method_detail}
This strategy aims to align the holistic semantic content of the sign language video (represented by pose features) with the corresponding text translation.

\textbf{1. Part-Specific Hyperbolic Embeddings}:
The temporally mean-pooled Euclidean feature vectors $\bar{\mathbf{f}}_p$ for each anatomical part $p$ (body, hands, face) are projected into the Poincaré ball $\mathbb{B}_{c}^{d_{\text{hyp}}}$. This projection, yielding hyperbolic embeddings $\mathbf{h}_p$, is achieved using the \texttt{HyperbolicProjection} layer (\Cref{lst:hyperbolic_projection} in \Cref{sec:code_listings}), as defined in Eq.~(4) of the main paper:
\begin{equation}\label{eq:supp_part_proj_pooled_final}
    \mathbf{h}_p = \exp_{\mathbf{0}}^c(s_p \mathbf{W}^{p} \bar{\mathbf{f}}_p).
\end{equation}
Here, $\mathbf{W}^{p}$ represents a linear layer for part $p$, and $s_p$ is a learnable scalar that adaptively scales the tangent space representation before the exponential map $\exp_{\mathbf{0}}^c(\cdot)$ projects it onto the manifold.

\textbf{2. Weighted Fréchet Mean for Global Pose Representation}:
The set of part-specific hyperbolic embeddings $\{\mathbf{h}_p\}$ is aggregated into a single global pose representation $\boldsymbol{\mu}_{\text{pose}} \in \mathbb{B}_{c}^{d_{\text{hyp}}}$. This is achieved by computing their weighted Fréchet mean, which is the hyperbolic analogue of a weighted average. The Fréchet mean is defined as the point that minimizes the sum of squared weighted geodesic distances to all input points:
\begin{align}
    \boldsymbol{\mu}_{\text{pose}} &= \underset{\boldsymbol{\mu} \in \mathbb{B}_{c}^{d_{\text{hyp}}}}{\operatorname{argmin}} \sum_{p=1}^P w_p d_{\mathbb{B}_c}^2(\boldsymbol{\mu}, \mathbf{h}_p). \label{eq:supp_frechet_def_pooled_final}
\end{align}
The weights $w_p$ are designed to give more importance to parts whose embeddings are further from the origin of the Poincaré ball (i.e., parts with more "hyperbolic energy" or distinctness), normalised via softmax:
\begin{equation}\label{eq:supp_frechet_weights_pooled_final}
    w_p = \frac{\exp(d_{\mathbb{B}_c}(\mathbf{0}, \mathbf{h}_p) / \lambda_w)}{\sum_{j=1}^P \exp(d_{\mathbb{B}_c}(\mathbf{0}, \mathbf{h}_j) / \lambda_w)}.
\end{equation}
Here, $\lambda_w$ is a temperature parameter for the softmax (e.g., fixed to 1.0 in our experiments) controlling the sharpness of the weight distribution. The computation is performed iteratively as detailed in Algorithm~1 of the main paper and \Cref{lst:frechet_mean} (\Cref{sec:code_listings}).

\textbf{3. Global Text Representation}:
Similarly, a global hyperbolic text embedding $\mathbf{h}_{\text{text}} \in \mathbb{B}_{c}^{d_{\text{hyp}}}$ is derived from the mT5 model's output. Euclidean token embeddings from the final layer of the mT5 decoder are first mean-pooled (respecting padding masks) to obtain a single sentence-level vector $\bar{\mathbf{e}}_{\text{text}}$. This vector is then projected into $\mathbb{B}_{c}^{d_{\text{hyp}}}$ using a dedicated hyperbolic projection layer (structurally identical to Eq.~\eqref{eq:supp_part_proj_pooled_final}):
\begin{equation}\label{eq:supp_text_proj_pooled_method_final}
    \mathbf{h}_{\text{text}} = \exp_{\mathbf{0}}^c(s_{\text{text}} \mathbf{W}^{\text{text}} \bar{\mathbf{e}}_{\text{text}}).
\end{equation}
The implementation details are shown in \Cref{lst:pooled_text_emb} (\Cref{sec:code_listings}).

\textbf{4. Contrastive Alignment}:
Finally, the geometric contrastive loss (Eq.~(5) in the main paper) is applied between batches of these global pose embeddings $\{\boldsymbol{\mu}_{\text{pose},i}\}$ and global text embeddings $\{\mathbf{h}_{\text{text},i}\}$. This encourages semantically similar pose-text pairs to be closer in hyperbolic space.

\subsubsection{Token Method (Fine-Grained Part-Text Alignment)}
\label{app:token_method_detail_final}
This strategy facilitates a more detailed alignment by relating individual pose part embeddings $\{\mathbf{h}_p\}$ with contextually relevant text segment embeddings $\{\mathbf{c}_p\}$.

\textbf{1. Hyperbolic Pose Part Embeddings $\{\mathbf{h}_p\}$}:
These are obtained exactly as in the Pooled Method, using Eq.~\eqref{eq:supp_part_proj_pooled_final}. Each $\mathbf{h}_p$ represents a specific anatomical part's overall kinematic signature.

\textbf{2. Hyperbolic Text Token Embeddings}:
Instead of a global text embedding, each Euclidean text token embedding $\mathbf{e}_{\text{token},j}$ (from the mT5 decoder's final layer) is individually projected into the Poincaré ball $\mathbb{B}_{c}^{d_{\text{hyp}}}$:
\begin{equation}\label{eq:supp_text_token_proj_method_final}
    \mathbf{h}_{\text{token},j} = \exp_{\mathbf{0}}^c(s_{\text{text}} \mathbf{W}^{\text{text}} \mathbf{e}_{\text{token},j}).
\end{equation}
This results in a sequence of hyperbolic token embeddings $\{\mathbf{h}_{\text{token},j}\}_{j=1}^{L_t}$, where $L_t$ is the text sequence length.

\textbf{3. Hyperbolic Attention Mechanism}:
For each hyperbolic pose part embedding $\mathbf{h}_p$ (acting as a query), a contextual text embedding $\mathbf{c}_p$ is generated. This is achieved using a hyperbolic attention mechanism (see \Cref{lst:token_attention} in \Cref{sec:code_listings}) that operates as follows:
    \begin{itemize}[noitemsep,topsep=3pt,parsep=3pt,partopsep=0pt]
        \item \textbf{Key Transformation}: The hyperbolic text token embeddings $\{\mathbf{h}_{\text{token},j}\}$ serve as keys. The embeddings are first transformed using learnable Möbius transformations to enhance their representational capacity:
            $$ \mathbf{k}_j = (\mathbf{M}_{\text{key}} \otimes_c \mathbf{h}_{\text{token},j}) \oplus_c \mathbf{b}_{\text{key}}, $$
            where $\mathbf{M}_{\text{key}}$ is a learnable Möbius matrix and $\mathbf{b}_{\text{key}}$ is a learnable Möbius bias vector.
        \item \textbf{Attention Scores}: Attention scores are computed based on the negative geodesic distance between each pose query $\mathbf{h}_p$ and each transformed text key $\mathbf{k}_j$:
            $$ \text{score}_{pj} = -d_{\mathbb{B}_c}(\mathbf{h}_p, \mathbf{k}_j). $$
        \item \textbf{Attention Weights}: These scores are normalised using a softmax function (after applying padding masks) to obtain attention weights $\alpha_{pj}$:
            $$ \alpha_{pj} = \text{softmax}\left(\frac{\text{score}_{pj}}{\tau_{\text{attn}}}\right), $$
            where $\tau_{\text{attn}}$ is a learnable temperature parameter for the attention mechanism, distinct from the temperature in the contrastive loss.
        \item \textbf{Contextual Text Embedding $\mathbf{c}_p$}: The contextual text embedding $\mathbf{c}_p$ corresponding to pose part $\mathbf{h}_p$ is then computed as the hyperbolic weighted midpoint of the original hyperbolic text token embeddings $\{\mathbf{h}_{\text{token},j}\}$, using the attention weights $\{\alpha_{pj}\}$.
    \end{itemize}

\textbf{4. Contrastive Alignment}:
The geometric contrastive loss (Eq.~(5), main paper) is then applied for each pair $(\mathbf{h}_{p,i}, \mathbf{c}_{p,i})$ across the batch. The total regularization loss for this strategy is the average of these individual contrastive losses over all parts $P$.

\subsubsection{Intuition Behind the Token Method}
\label{app:token_method_intuition_final}
While the Pooled Method aligns the overall semantics of a sign sequence with its translation, it may not capture how specific signing elements (e.g., a handshape, movement, or facial expression) correspond to particular words or phrases. The Token Method aims to establish this more fine-grained understanding.

The core intuition is as follows:
\begin{enumerate}[noitemsep,topsep=3pt,parsep=3pt,partopsep=0pt]
    \item \textbf{Compositional Language Understanding}: Sign languages, like spoken/written languages, are compositional. Different articulators (hands, body, face) convey distinct lexical or grammatical information. The Token Method attempts to map these compositional units from pose to corresponding textual tokens (words/sub-words).

    \item \textbf{Targeted Part-to-Segment Alignment}: Instead of a single global comparison, this method learns to connect individual pose part representations (e.g., features for the dominant hand, to the most relevant segments of the textual translation.

    \item \textbf{Pose Parts as Queries, Text Tokens as Sources}: Each hyperbolic pose part embedding $\mathbf{h}_p$ acts as a "query", effectively asking: "Which text tokens are most semantically relevant to this pose feature?" The sequence of hyperbolic text token embeddings $\{\mathbf{h}_{\text{token},j}\}$ serves as the ``information source'' for these queries.

    \item \textbf{Hyperbolic Attention for Geometric Relevance}:
        \begin{itemize}[noitemsep,topsep=2pt,parsep=2pt,partopsep=0pt]
            \item Relevance between a pose part query $\mathbf{h}_p$ and a (transformed) text token key $\mathbf{k}_j$ is measured by their geodesic distance $d_{\mathbb{B}_c}(\mathbf{h}_p, \mathbf{k}_j)$ in the learned hyperbolic space. A smaller distance implies higher relevance. Using hyperbolic geometry allows these comparisons to potentially leverage latent hierarchical relationships between concepts.
            \item Learnable Möbius transformations on text tokens (to get keys $\mathbf{k}_j$) enable the model to learn distinct tokens relevant to different pose parts (e.g., a verb token might be transformed to be closer to a body movement embedding).
            \item The attention weights $\alpha_{pj}$ then quantify the contribution of each text token $j$ to the meaning conveyed by pose part $p$.
        \end{itemize}

    \item \textbf{Learning Textual Context for Each Pose Part}: The contextual text embedding $\mathbf{c}_p$ is a hyperbolic weighted midpoint of all text token embeddings, using the attention weights $\alpha_{pj}$. Thus, $\mathbf{c}_p$ is a summary of the sentence, but specifically customised by the interaction of pose part $p$.

    \item \textbf{Refined Contrastive Learning}: The model is regularised to make each pose part embedding $\mathbf{h}_p$ close to its corresponding contextual text view $\mathbf{c}_p$ in hyperbolic space, while pushing it away from non-corresponding pairs.

    \item \textbf{Overall Benefit}: This detailed, part-specific alignment encourages the mT5 model to learn more precise mappings between kinematic features of different articulators and semantic units within the text. For example, it can help distinguish visually similar signs based on subtle hand details (encoded in $\mathbf{h}_{\text{hand}}$) that correlate with specific words, leading to more accurate and nuanced translations.
\end{enumerate}

\section{Mathematical Foundations }
\label{sec:supp_math}

This section recalls two geometric components that Geo-Sign relies on:

\begin{itemize}[noitemsep,topsep=1pt,parsep=1pt]
    \item the \emph{Weighted Fréchet Mean} inside the Poincaré ball
          (used in Algorithm 1 of the paper);
    \item the Euclidean gradient of the hyperbolic distance that appears
          in the contrastive loss.
\end{itemize}

\subsection{Fréchet Mean in the Poincaré Ball}
\label{sec:supp_frechet}

Given points $x_1,\dots,x_N$ in a metric space $(\mathcal M,d)$ with
normalised weights $w_i>0$, $\sum_i w_i=1$, the
\textbf{Fréchet mean} minimises
\[
  \mathcal F(\mu)=\sum_{i=1}^N w_i\,d^2(\mu,x_i),
  \qquad
  \mu^\star=\arg\min_{\mu\in\mathcal M}\mathcal F(\mu).
\]

\medskip\noindent\textbf{Why not simply average the embeddings in Euclidean space?}
Two issues appear inside the curved Poincaré ball:

\begin{enumerate}[label=(\alph*), leftmargin=2.3em,itemsep=2pt]
  \item \textbf{Manifold constraint.}  
        A Euclidean average of interior points can fall \emph{outside}
        the ball, i.e.\ outside valid hyperbolic space, forcing an
        ad-hoc projection that distorts geometry.
  \item \textbf{Metric distortion.}  
        Euclidean distance underestimates separation near the boundary
        because the hyperbolic metric stretches space there.
        A straight average therefore over-emphasises central points and
        washes out fine structure carried by peripheral ones.
\end{enumerate}

The intrinsic Fréchet mean lives on the manifold and uses the true
hyperbolic distance, so it respects curvature.

\medskip\noindent\textbf{Why distance-based weights?}\;
Each pose part (body, face, left hand, right hand) yields a hyperbolic
embedding $h_p$.
We set
\(
  w_p \propto
  \exp\!\bigl(d_{\mathbb B_c}(0,h_p)/\lambda_w\bigr)
\)
so parts farther from the origin, in regions of higher curvature and
greater discriminative power, receive more influence.
Without this weighting the mean would drift toward the centre, diluting
information contributed by the hands and face.

\paragraph{Iterative update.}
On any Riemannian manifold the mean is found by Riemannian
gradient descent; the update at iteration $k$ is
\begin{equation}
  \mu^{(k+1)}
  =\exp_{\mu^{(k)}}\!\Bigl(
        \eta_k\sum_{i=1}^N w_i\,
        \log_{\mu^{(k)}}(x_i)
      \Bigr),
\label{eq:supp_frechet_update}
\end{equation}
with step size $\eta_k>0$.

\begin{proposition}[Convergence in $\mathbb B_c^{d}$]
\label{prop:supp_frechet}
The Poincaré ball $\mathbb B_c^{d}$ is a Hadamard manifold, hence
$\mathcal F$ is strictly convex and has a \emph{unique} minimiser
$\mu^\star$.
Let $L$ be the Lipschitz constant of
$\nabla\mathcal F$ on the geodesic convex hull of $\{x_i\}$.
If $0<\eta_k\le 2/L$ for all $k$, the iterates
\eqref{eq:supp_frechet_update} converge to $\mu^\star$.
In practice we observe $L\!\le\!2$, so the simple choice
$\eta_k=1$ is usually sufficient and used in our approach.
\end{proposition}

\subsection{Gradient of the Hyperbolic Distance}
\label{sec:supp_grad}

For $u,v\in\mathbb B_c^{d}$ let $w=(-u)\oplus_c v$
(the Möbius difference, i.e.\ the "vector" from $u$ to $v$ transported
to the origin).
The Poincaré distance is
\[
  d_{\mathbb B_c}(u,v)
  =\frac{2}{\sqrt c}\,
   \operatorname{artanh}\bigl(\sqrt c\,\|w\|_2\bigr).
\]

Differentiating
\cite{nickel2017poincare,ganea2018hyperbolic} gives the Euclidean
gradient required for autograd:
\begin{equation}
  \nabla_{u}\,d_{\mathbb B_c}(u,v)
  =-\frac{2}{\lambda_u^c\,\lambda_v^c}\,
    \frac{w}{\|w\|_2}\,
    \frac{1}{1-c\|w\|_2^{\,2}}
\label{eq:supp_grad_distance}
\end{equation}
with conformal factor
\(
  \displaystyle
  \lambda_x^{c}=\frac{2}{1-c\|x\|_2^{2}}.
\)
The same formula (with sign reversed) holds for $\nabla_{v}$.

\medskip\noindent
The update rule
\eqref{eq:supp_frechet_update} and the gradient
\eqref{eq:supp_grad_distance} provide all the geometric tools
needed by Geo-Sign’s hyperbolic contrastive regulariser.

\section[Learnable Model Parameters]{Learnable Model Parameters: $c$ and \texorpdfstring{$\alpha$}{alpha}}
\label{app:learnable_model_params}
Our Geo-Sign model incorporates several learnable parameters beyond standard network weights. This section details two key ones: the manifold curvature $c$ and the loss blending factor $\alpha$.

\subsection{Discussion on Learnable Curvature}
The curvature of the Poincaré ball, $\kappa = -c$ (where $c > 0$), is a crucial hyperparameter that dictates the ``shape'' of the hyperbolic space. Instead of fixing $c$ heuristically, we make it a learnable parameter of our model (see \Cref{lst:manifold_init} in \Cref{sec:code_listings}).

\textbf{Optimization Strategy}:
The curvature magnitude $c$ is initialised (e.g., via \texttt{args.init\_c} as mentioned in the main paper's experiments) and then updated via standard gradient descent as part of the end-to-end training process. The \texttt{geoopt} library facilitates this by defining $c$ as an \texttt{nn.Parameter} within its \texttt{PoincareBall} manifold object when \texttt{learnable=True}.

\begin{figure}[htbp]
  \centering
  \begin{subfigure}[b]{0.48\linewidth}
    \centering
    \includegraphics[width=\linewidth]{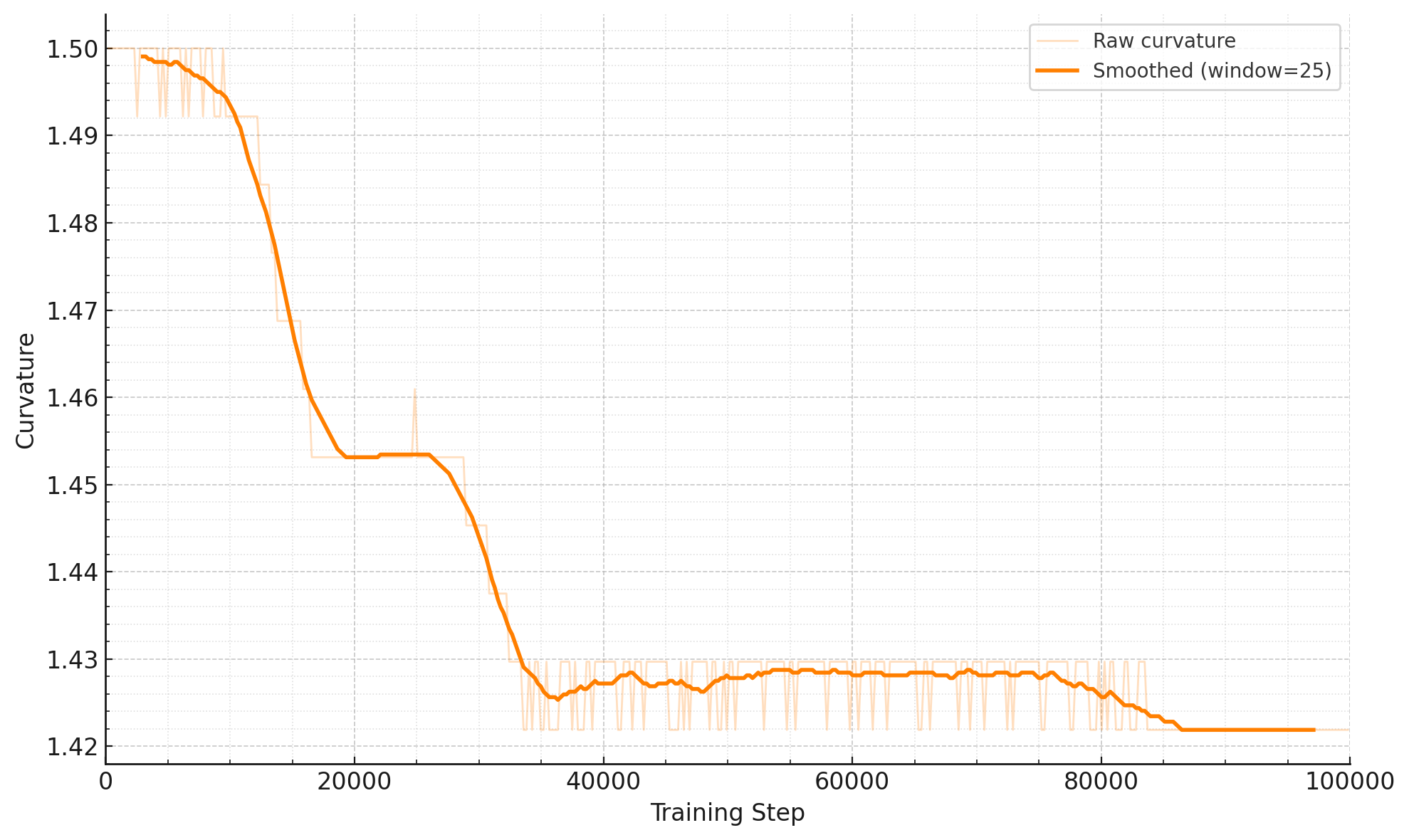}
    \caption{$c$ initialised at 1.50, decreases and stabilizes around 1.42.}
    \label{fig:curvature_a}
  \end{subfigure}
  \hfill
  \begin{subfigure}[b]{0.48\linewidth}
    \centering
    \includegraphics[width=\linewidth]{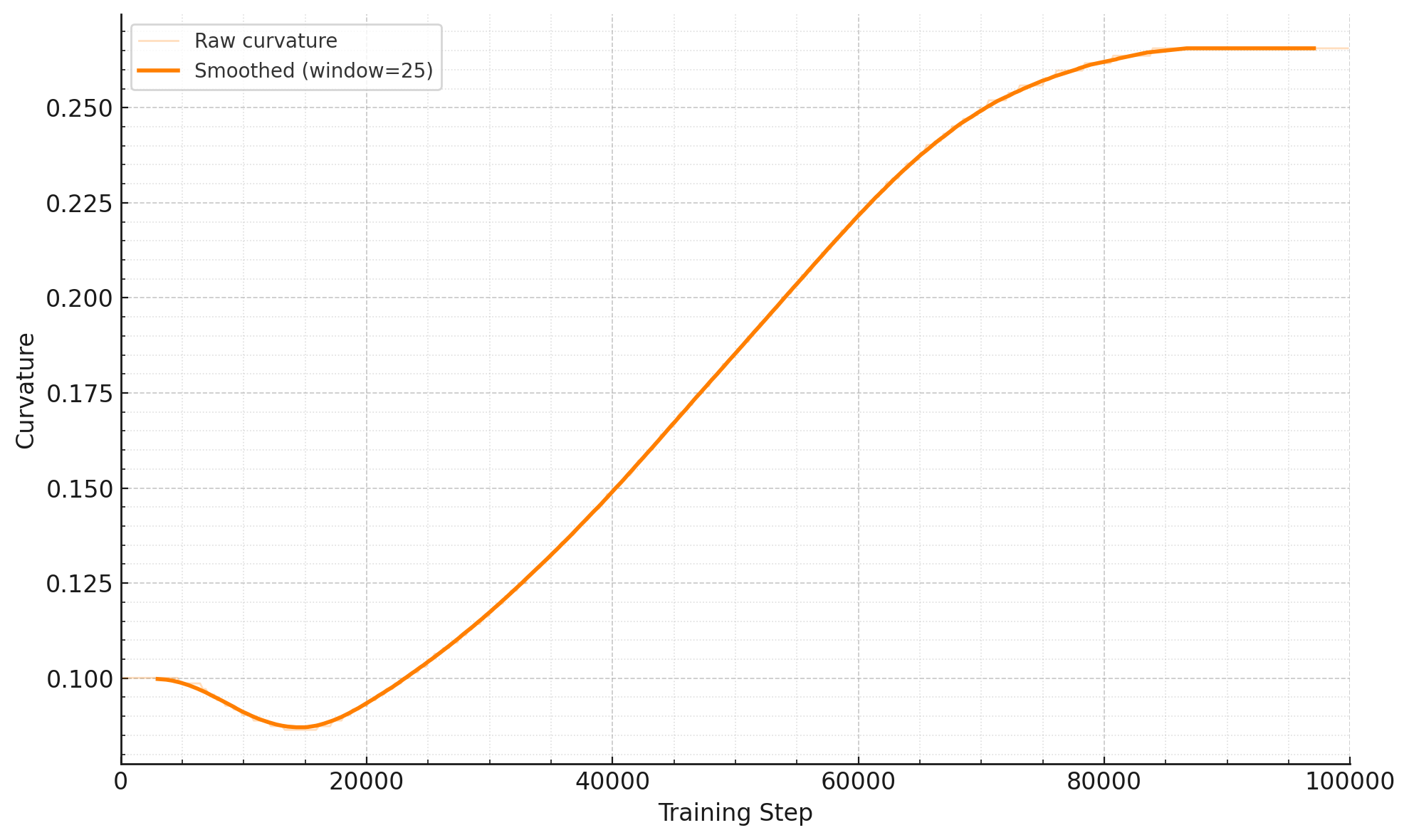}
    \caption{$c$ initialised at 0.10, increases and stabilizes around 0.20.}
    \label{fig:curvature_b}
  \end{subfigure}
  \caption{Evolution of the learnable manifold curvature $c$ during training for different initializations.
           (\subref{fig:curvature_a}) When initialised at $c=1.50$, the curvature magnitude slightly decreases, suggesting an optimal value around $1.42$ for this setup.
           (\subref{fig:curvature_b}) When initialised at a low $c=0.10$, the curvature increases, indicating the model benefits from more ``hyperbolic space'' initially. It stabilizes around $c=0.20$, potentially influenced by the dynamic $\alpha$ schedule that reduces regularization emphasis over time.}
  \label{fig:curvature_side_by_side}
\end{figure}

The main paper's ablation studies (Table 2a) show that initializing $c$ in the range of $1.0-2.0$ (e.g., optimal BLEU-4 at $c=1.5$) yields strong performance. \Cref{fig:curvature_side_by_side} illustrates how $c$ adapts during training from different initializations.

\subsection[Discussion on Loss Blending Factor alpha]{Discussion on Loss Blending Factor \texorpdfstring{$\alpha$}{alpha}}
The total training loss $\mathcal{L}_{\text{total}}$ is a weighted combination of the primary cross-entropy translation loss $\mathcal{L}_{\text{CE}}$ and our hyperbolic contrastive regularization term $\mathcal{L}_{\text{hyp\_reg}}$:
$$ \mathcal{L}_{\text{total}} = \alpha \cdot \mathcal{L}_{\text{CE}} + (1-\alpha) \cdot \mathcal{L}_{\text{hyp\_reg}}. $$
The blending factor $\alpha$ is not fixed but is dynamically adjusted during training. This dynamic scheduling allows the model to potentially benefit from different loss emphases at different training stages. The calculation of $\alpha$ at each training step (see \Cref{lst:alpha_calc} in \Cref{sec:code_listings}) is:
\begin{equation}\label{eq:supp_alpha_schedule_math_final}
    \alpha_{\text{final}} = \text{clamp} \left( (\alpha_{\text{init}} + 0.1 \cdot \text{progress}) + \sigma(\text{logit}_{\alpha}) \cdot 0.2, \; 0.1, \; 1.0 \right),
\end{equation}
where:
\begin{itemize}[noitemsep,topsep=3pt,parsep=3pt,partopsep=0pt]
    \item $\alpha_{\text{init}}$ is the initial value for the blending factor, specified as a hyperparameter (e.g., \texttt{args.alpha = 0.7} from the main paper's ablations, Table 2b, which was found to be optimal).
    \item $\text{progress}$ is the current training progress, calculated as $\frac{\text{current\_training\_step}}{\text{total\_training\_steps}}$, ranging from 0 to 1. This component introduces a linear ramp, potentially increasing $\alpha$'s baseline by up to 0.1 over the course of training.
    \item $\text{logit}_{\alpha}$ is an \texttt{nn.Parameter} (a learnable scalar, referred to as \texttt{self.loss\_alpha\_logit} in the code). $\sigma(\cdot)$ is the sigmoid function, so $\sigma(\text{logit}_{\alpha})$ maps this learnable scalar to the range $(0,1)$. This term provides a learnable adjustment to $\alpha$ in the range of $[0, 0.2]$.
    \item $\text{clamp}(\cdot, 0.1, 1.0)$ ensures that the final $\alpha_{\text{final}}$ remains within the bounds $[0.1, 1.0]$.
\end{itemize}
This dynamic $\alpha$ allows for an initial phase where the hyperbolic regularization might have more relative influence (if $\alpha_{\text{init}}$ is smaller), gradually shifting emphasis or allowing the model to fine-tune the balance via the learnable component. The ablation study in the main paper (Table 2b) indicates that an initial $\alpha_{\text{init}}=0.7$ (i.e., 30\% weight to $\mathcal{L}_{\text{hyp\_reg}}$ initially) provides the best results, highlighting the complementary role of the hyperbolic regularization.

\begin{figure}[htbp]
    \centering
    \includegraphics[width=0.9\linewidth]{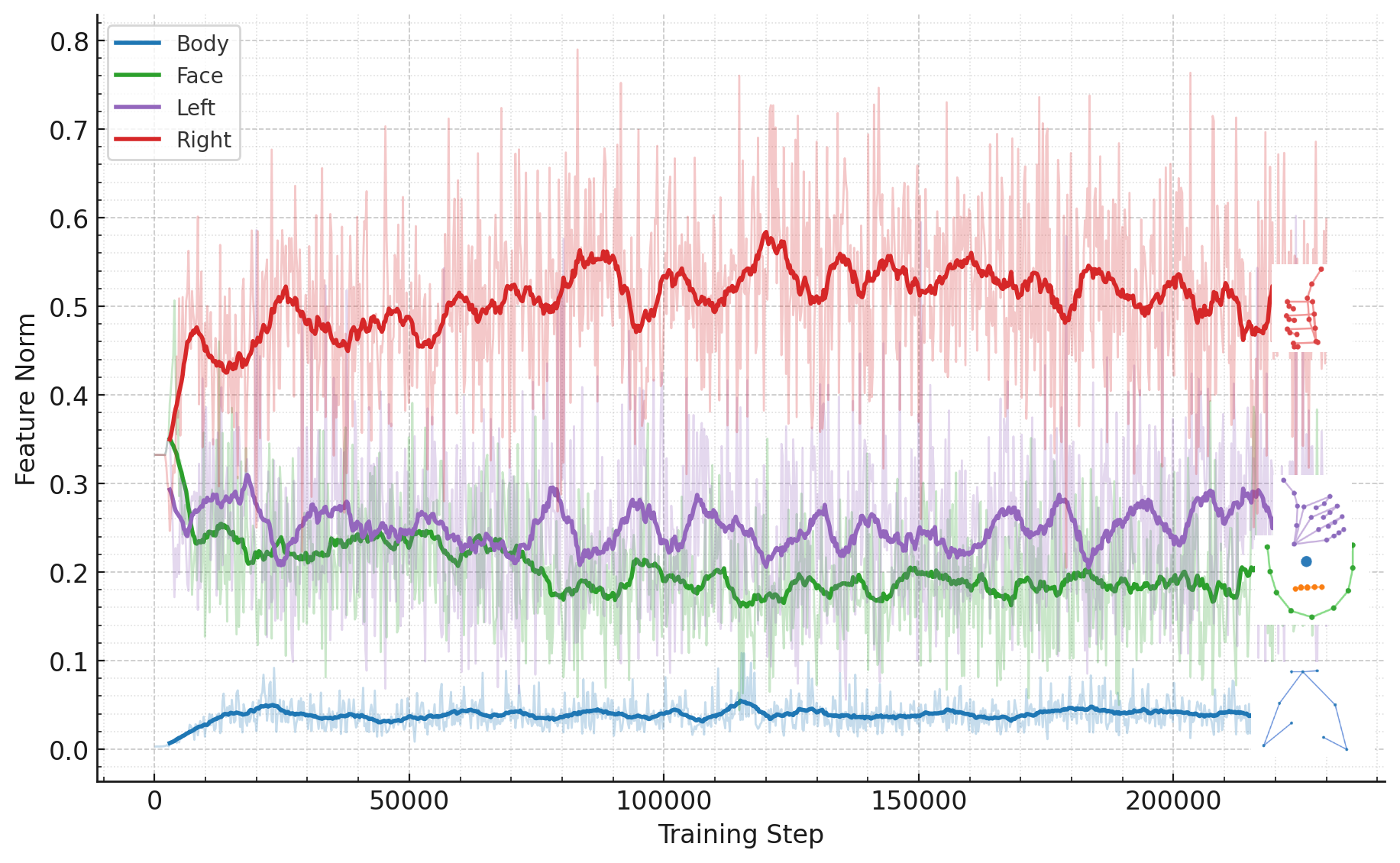}
    \caption{Plot of the geodesic distances from the origin ($\mathbf{0}$) of the Poincaré disk to the hyperbolic pose embeddings ($\mathbf{h}_p$) during training, averaged per part type. This shows how features for different parts utilize the hyperbolic space. For instance, right hand features (often conveying detailed lexical information) tend to move further from the origin, leveraging more of the hyperbolic curvature for discriminability. Body and face features, which might represent broader semantics or prosody, may remain closer to the Euclidean-like central region.}
    \label{fig:geodesic_distances_plot}
\end{figure}

\begin{figure}[htbp]
    \centering
    \includegraphics[width=0.9\linewidth]{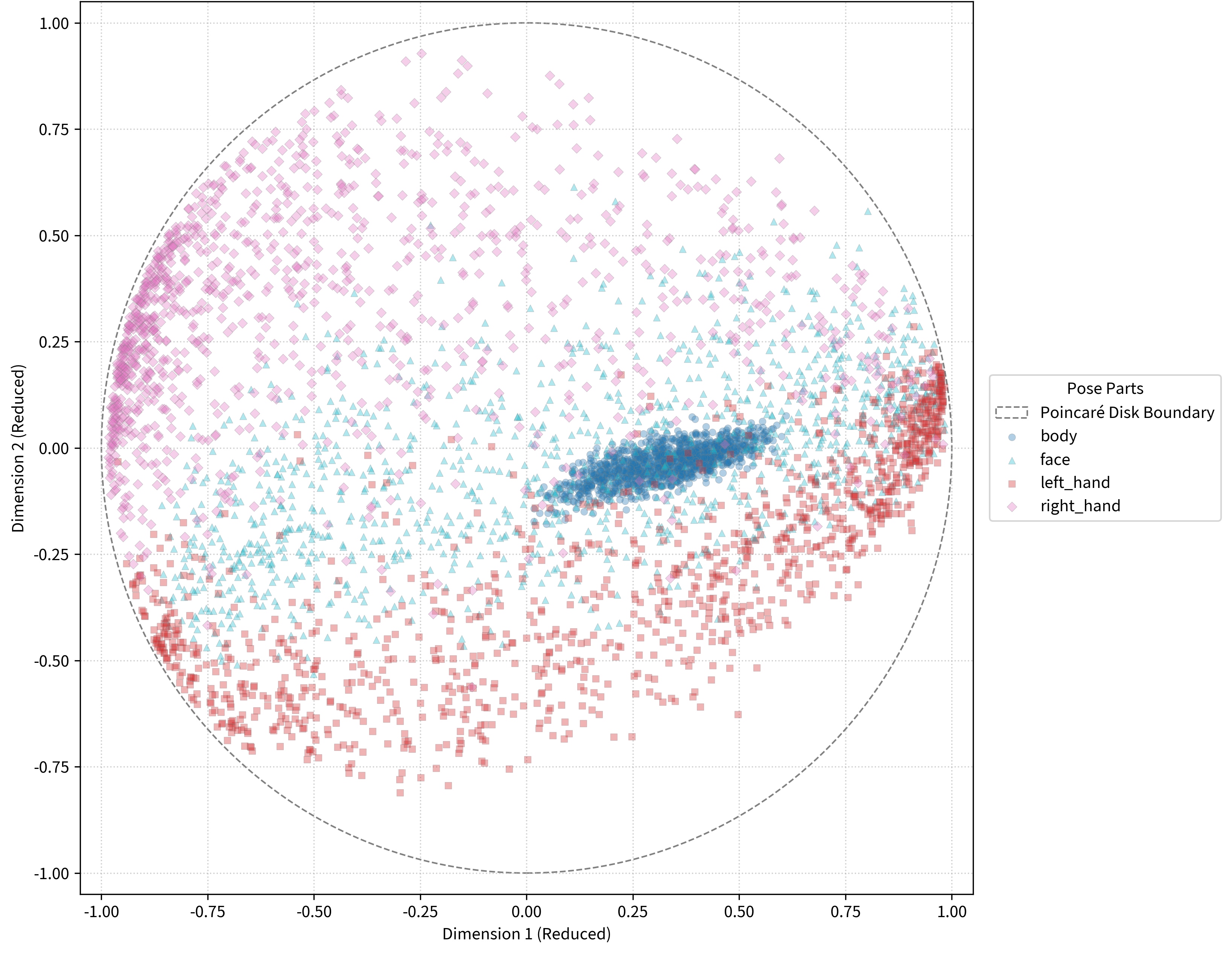}
    \caption{PCA projection of 1000 hyperbolic pose part embeddings (log-mapped to the tangent space at origin, then PCA-reduced to 2D) visualised within the Poincaré disk. Body features (blue) are tightly clustered near the origin, suggesting their discriminability is well-handled in a more Euclidean-like region. Hand features (left: red square, right: pink diamond) and face features (light blue triangle) are more dispersed, with hand features often pushed towards the periphery. This indicates these parts benefit from the increased representational capacity near the boundary of the Poincaré disk, where hyperbolic geometry provides more ``space'' to distinguish subtle variations crucial for sign language semantics.}
    \label{fig:poincare_projection_plot}
\end{figure}

\section{Experimental Setup, Analysis, and Qualitative Results}
\label{app:exp_setup_analysis}

\subsection[Computational Considerations and Profiler Analysis]{Computational Profile}
\label{app:computational_profile}
This section discusses the computational profile of Geo-Sign, comparing it to a baseline Uni-Sign (Pose) model without hyperbolic regularization. The analysis is based on DeepSpeed profiler outputs for models run with a batch size of 8 on the CSL-Daily dataset for the Sign Language Translation (SLT) task.

\textbf{Experimental Context}:
Key experimental conditions for fine-tuning include:
\begin{itemize}[noitemsep,topsep=3pt,parsep=3pt,partopsep=0pt]
    \item \textbf{Hardware}: 4 NVIDIA RTX 3090 GPUs.
    \item \textbf{Training Time}: Approximately 10 hours for 40 epochs of fine-tuning on CSL-Daily.
    \item \textbf{Precision}: Mixed-precision training (\texttt{bfloat16}) is used for standard PyTorch layers, while \texttt{float32} is maintained for Geoopt hyperbolic operations to ensure numerical stability.
    \item \textbf{Batching Strategy}: With an effective batch size of 8 per GPU, the model occupies $\approx$20GB of memory. During training, we increase the total batch size to 32 and accumulate gradients over 8 steps, achieving a hypothetical batch size of 256. For the following profiler analysis, we report results for a single GPU with a batch size of 8 to provide a clear per-device profile.
\end{itemize}

\subsubsection{Profiler Summary and Comparative Analysis}
\Cref{tab:profiler_comparison_bs8_final_app_revised_again} summarizes key metrics from the profiler. Parameter counts are consistent with the main paper's Table~1, while MACs (Multiply-Accumulate operations) and Latency are derived from DeepSpeed profiler outputs for a batch size of 8. \Cref{tab:csl_daily_params_compare_final} provides a comparison of model parameters against other gloss-free methods.

\begin{table*}[htbp]
\centering
\caption{Computational profile comparison at Batch Size 8: Baseline Uni-Sign (Pose) vs. Geo-Sign variants. Parameter counts from main paper's Table~1. MACs and Latency from DeepSpeed profiler outputs. ``Hyperbolic Proj. Layer MACs'' reflects profiled contributions from the learnable linear transformations within these layers.}
\label{tab:profiler_comparison_bs8_final_app_revised_again}
\sisetup{round-mode=places,round-precision=2,table-format=3.2,tight-spacing=true}
\resizebox{\textwidth}{!}{%
\begin{tabular}{l S S S S[table-format=2.2] S S[table-format=3.1]}
\toprule
\textbf{Model Variant (Batch Size 8)} & {\textbf{Total Params (M)}} & {\textbf{Added Params (M)}} & {\textbf{Total Fwd MACs (GMACs)}} & {\textbf{Hyperbolic Proj. Layer MACs (MMACs)}} & {\textbf{Fwd Latency (ms)}} & {\textbf{Latency Increase (\%)}} \\
\midrule
Baseline Uni-Sign (Pose) & 587.75 & {-} & 116.59 & {-} & 415.73 & {-} \\
Geo-Sign (Hyperbolic Pooled) & 588.21 & 0.46 & 116.60 & 3.67 & 1630.00 & 292.1 \\
Geo-Sign (Hyperbolic Token) & 589.10 & 1.35 & 116.60 & {\text{$\approx$}9.96} & 2550.00 & 513.4 \\
\bottomrule
\end{tabular}%
}
\end{table*}

\begin{table*}[t]
\centering
\caption{Sign Language Translation performance (Test Set: BLEU-4, ROUGE-L) and model parameters on CSL-Daily. Scores are percentages (\%). Higher is better. `Pose' and `RGB' indicate input modalities. VE/LM/Total Params are in Millions (M). Approx. values indicated by $\approx$. Data from CSL-Daily (Train: 18,401 sentences / 20.62 hours).}
\label{tab:csl_daily_params_compare_final}
\resizebox{\textwidth}{!}{%
\begin{tabular}{l >{\raggedright\arraybackslash}p{2.5cm} >{\raggedright\arraybackslash}p{1.1cm} >{\raggedright\arraybackslash}p{2.8cm} >{\raggedright\arraybackslash}p{1.3cm} >{\raggedright\arraybackslash}p{1.1cm} |cc|cc}
\toprule
\multirow{2}{*}{\textbf{Method}} & \multirow{2}{*}{\textbf{VE Name}} & \textbf{VE} & \multirow{2}{*}{\textbf{LM Name}} & \textbf{LM} & \textbf{Total} & \multicolumn{2}{c|}{\textbf{Modality}} & \multicolumn{2}{c}{\textbf{Test Set}} \\
\cmidrule(lr){7-8}\cmidrule(lr){9-10}
& & \textbf{Params} & & \textbf{Params} & \textbf{Params} & Pose & RGB & B-4 & R-L \\
& & \textbf{(M)} & & \textbf{(M)} & \textbf{(M)} & & & & \\

\midrule
\rowcolor[gray]{.9}\multicolumn{10}{c}{\textit{Gloss-Free Methods (Prior Art)}} \\
MSLU~\cite{zhou2024scaling} & EffNet& 5.3 & mT5-Base & 582.4 & 587.7 & \checkmark & -- & 11.42 & 33.80 \\
SLRT~\cite{camgoz2020sign} (G-Free) & EffNet & 5.3 & Transformer & $\approx$30 & $\approx$35.3 & -- & \checkmark & 3.03 & 19.67 \\
GASLT~\cite{yin2023gloss} & I3D & 13 & Transformer & $\approx$30 & $\approx$43.0 & -- & \checkmark & 4.07 & 20.35 \\
GFSLT-VLP~\cite{zhou2023gloss} & ResNet18 & 11.7 & mBart & 680 & 691.7 & -- & \checkmark & 11.00 & 36.44 \\
FLa-LLM~\cite{chen2024factorized} & ResNet18 & 11.7 & mBart & 680 & 691.7 & -- & \checkmark & 14.20 & 37.25 \\
Sign2GPT~\cite{wong2024sign2gpt} & DinoV2 & 21.0 & XGLM & 1732.9 & 1753.9 & -- & \checkmark & 15.40 & 42.36 \\
SignLLM~\cite{gong2024signllm} & ResNet18 & 11.7 & LLaMA-7B & 6738.4 & 6750.1 & -- & \checkmark & 15.75 & 39.91 \\
C$^{2}$RL~\cite{chen2024c2rl} & ResNet18 & 11.7 & mBart & 680 & 691.7 & -- & \checkmark & 21.61 & 48.21 \\
\midrule
\rowcolor[gray]{.9}\multicolumn{10}{c}{\textit{Our Models and Baselines}} \\
Uni-Sign~\cite{li2025uni} (Pose) & GCN & 5.3 & mT5-Base & 582.4 & 587.7 & \checkmark & -- & 25.61 & 54.92 \\
Uni-Sign~\cite{li2025uni} (Pose+RGB) & EffNet+GCN & 9.7 & mT5-Base & 582.4 & 592.1 & \checkmark & \checkmark & 26.36 & 56.51 \\
\midrule
Geo-Sign (Hyperbolic Pooled) & GCN+Geo & 5.8 & mT5-Base & 582.4 & 588.21 & \checkmark & -- & 27.17 & 57.75 \\
\textbf{Geo-Sign (Hyperbolic Token)} & GCN+Geo+Attn & 6.7 & mT5-Base & 582.4 & 589.1 & \checkmark & -- & \textbf{27.42} & \textbf{57.95} \\
\bottomrule
\end{tabular}
}
\end{table*}

\textbf{Parameter Overhead}:
The increase in parameters due to the hyperbolic components is marginal compared to the overall model size, which is dominated by the mT5 language model ($\approx 582.4\text{M}$ parameters).
\begin{itemize}[noitemsep,topsep=3pt,parsep=3pt,partopsep=0pt]
    \item Baseline Uni-Sign (Pose): $\approx 587.75\text{M}$ parameters.
    \item \textbf{Geo-Sign (Pooled)}: Adds $\approx 0.46\text{M}$ parameters, primarily from the five hyperbolic projection layers (one for each of the four pose parts and one for the pooled text embedding).
    \item \textbf{Geo-Sign (Token)}: Adds $\approx 1.35\text{M}$ parameters. This includes the $\approx 0.46\text{M}$ for projection layers plus an additional $\approx 0.89\text{M}$ for the learnable parameters within the hyperbolic attention mechanism (Möbius matrices and biases).
\end{itemize}
In both Geo-Sign variants, the parameter overhead from hyperbolic components is less than $0.25\%$ of the total model size. As shown in \Cref{tab:csl_daily_params_compare_final}, our Geo-Sign models achieve competitive or superior performance to recent RGB-based methods while maintaining a significantly smaller total parameter count. This highlights the efficiency of enhancing skeletal representations with geometric priors, challenging the trend that relies solely on scaling up visual encoders and language model decoders for performance gains in SLT.

\textbf{MACs Analysis}:
The DeepSpeed profiler indicates that the total forward MACs are very similar across all configurations at this batch size:
\begin{itemize}[noitemsep,topsep=3pt,parsep=3pt,partopsep=0pt]
    \item Baseline Uni-Sign (Pose): $\approx 116.59$ GMACs.
    \item Geo-Sign (Hyperbolic Pooled): $\approx 116.60$ GMACs. The profiler attributes $\approx 3.67$ MMACs to the linear transformations within its \texttt{HyperbolicProjection} layers.
    \item Geo-Sign (Hyperbolic Token): $\approx 116.60$ GMACs. Its \texttt{HyperbolicProjection} layers account for $\approx 9.96$ MMACs from their linear components.
\end{itemize}
The MACs from the learnable linear transformations within the hyperbolic projection layers constitute a very small fraction ($<0.01\%$) of the total model MACs. The bulk of MACs originates from the mT5 model (profiled at $\approx 66.29$ GMACs) and the ST-GCN modules (profiled at $\approx 49.93$ GMACs).
We should note, however, that standard profilers (like DeepSpeed's MAC counter) primarily quantify MACs from common operations like convolutions and linear layers. The computational cost of specialised geometric functions within \texttt{geoopt} (e.g., \texttt{manifold.dist}, \texttt{expmap0}, \texttt{logmap0}, Möbius arithmetic) is not explicitly broken out as distinct hyperbolic operation MACs. These functions often involve sequences of elementary operations that are not all MAC-based (e.g., square roots, divisions, trigonometric functions like \texttt{artanh} or \texttt{tanh}). Thus, their computational load may be underestimated by MAC counters and is often better reflected in measured latency.

\textbf{Latency Analysis}:
Latency figures clearly reveal the primary computational overhead introduced by the hyperbolic components during training:
\begin{itemize}[noitemsep,topsep=3pt,parsep=3pt,partopsep=0pt]
    \item Baseline Uni-Sign (Pose): $\approx 416 \text{ ms}$ forward latency per batch.
    \item Geo-Sign (Hyperbolic Pooled): $\approx 1630 \text{ ms}$ (1.63 s), an increase of $\approx 1214 \text{ ms}$ or $\approx 292\%$ over the baseline (approx. $3.9 \times$ slowdown).
    \item Geo-Sign (Hyperbolic Token): $\approx 2550 \text{ ms}$ (2.55 s), an increase of $\approx 2134 \text{ ms}$ or $\approx 513\%$ over the baseline (approx. $6.1 \times$ slowdown).
\end{itemize}
The substantial increase in training latency, despite modest increases in parameters and profiled MACs from learnable layers, underscores that the geometric operations themselves are the main performance consideration during the training phase. These operations (e.g., geodesic distance, exponential/logarithmic maps, Möbius transformations) are inherently more complex than their Euclidean counterparts.
The Token method is notably slower than the Pooled method during training due to its per-token hyperbolic attention.

Importantly, a key advantage of our regularization approach is that these geometric operations and the hyperbolic branch are \textbf{not utilised at inference time}. Consequently, Geo-Sign models incur no additional latency increase over the baseline Uni-Sign (Pose) model during inference, preserving efficiency for deployment.

\subsubsection{Discussion on Data Efficiency}
While not directly evaluated, it is hypothesised that skeletal data's abstraction from visual noise (lighting, background, clothing) can enhance robustness and generalization \cite{wong2025signrep}, especially when training data is limited. Hyperbolic geometry further imposes a structural prior on the representation space. This inductive bias could potentially improve data efficiency by guiding the learning process, particularly in scenarios with sparse data, although specific experiments to quantify this effect were not part of the current study. One trade-off of this approach is that we cannot directly leverage large pre-trained visual encoders as in the case of other RGB approaches, and so pre-training on a sign-specific dataset like CSL-News (1,985 hours, used by Uni-Sign) is essential. However, this pre-training data size is comparable to that used by other SLT methods which use datasets such as How2Sign \cite{duarte2021how2sign} (2000 hours) or YouTube-ASL \cite{uthus2024ytasl, tanzer2024ytfull} (6000 hours). We anticipate that our method would continue to scale well with larger pre-training datasets in other sign languages, though resource constraints prevented evaluation of this aspect.

\subsection[Further Technical Implementation Details]{Further Technical Implementation Details}
\label{app:technical_details_final}
This section provides additional details that are pertinent for a full understanding and potential reimplementation of Geo-Sign.
\begin{itemize}[noitemsep,topsep=3pt,parsep=3pt,partopsep=0pt]
    \item \textbf{Core Libraries}: Our implementation relies on PyTorch \cite{paszke2019pytorch} as the primary deep learning framework. For Transformer models, we utilize the HuggingFace Transformers library. All hyperbolic geometry operations and Riemannian optimization are handled by the Geoopt library \cite{geoopt2020kochurov}. For distributed training and profiling, DeepSpeed is employed.
    \item \textbf{Hyperparameter Tuning Strategy}: Key hyperparameters specific to the hyperbolic components, such as the initial curvature $c$, the initial loss blending factor $\alpha_{\text{init}}$ (referred to as \texttt{args.alpha} in code/main paper), and the hyperbolic embedding dimension $d_{\text{hyp}}$, were tuned using a grid search strategy on the CSL-Daily development set. Full hyperparameters are outlined in \cref{tab:hyperparameter_summary}.
    \item \textbf{Numerical Stability Measures}:
        \begin{itemize}[noitemsep,topsep=3pt,parsep=3pt,partopsep=0pt]
            \item Operations within \texttt{geoopt} are performed using \texttt{float32} precision to maintain numerical stability, while the rest of the model uses mixed precision.
            \item Small epsilon values (e.g., $10^{-5}$) are added in denominators and inside logarithms/arctanh functions where appropriate to prevent division by zeros.
            \item \textbf{Tangent Vector Clipping}: Before applying an exponential map $\exp_{\mathbf{x}}^c(\mathbf{v})$ from a point $\mathbf{x}$ with a tangent vector $\mathbf{v}$, especially $\exp_{\mathbf{0}}^c(\mathbf{v})$, it's crucial to ensure the resulting point remains strictly within the Poincaré ball and that the norm of $\mathbf{v}$ doesn't cause numerical issues in $\tanh(\cdot)$. We apply a clipping strategy as mentioned in Section 3.4 of the main paper:
                $$ \mathbf{v}_{\text{clipped}} \leftarrow \frac{\mathbf{v}}{\max(1, \sqrt{c}\|\mathbf{v}\|_2 + \epsilon_{\text{clip}})}, $$
                for a small $\epsilon_{\text{clip}} > 0$ (e.g., $10^{-5}$). This ensures that the argument to $\tanh$ in $\exp_{\mathbf{0}}^c$ does not become excessively large and that mapped points do not reach or exceed the boundary of the Poincaré ball. The \texttt{project=True} flag in \texttt{geoopt}'s \texttt{expmap} functions also helps enforce this by projecting points back onto the ball if they numerically fall outside.
        \end{itemize}
    \item \textbf{Gradient Clipping}: Standard norm-based gradient clipping is applied to all model parameters during training to stabilize the optimization process.
\end{itemize}

In \cref{tab:hyperparameter_summary} we provide the full hyper-parameters for the best performing model. The full code will be released following the review process.

\begin{table*}[htbp]
\centering
\caption{Hyperparameter summary for Geo-Sign experiments. Values are for the best reported model configuration.}
\label{tab:hyperparameter_summary}
\resizebox{\textwidth}{!}{%
\begin{tabular}{llll}
\toprule
\textbf{Category} & \textbf{Hyperparameter} & \textbf{Value} & \textbf{Description} \\
\midrule
\multicolumn{4}{l}{\textit{\textbf{General Training Configuration}}} \\
& Random Seed & 42 & Seed for reproducibility \\
& Training Epochs & 40 & Number of fine-tuning epochs on CSL-Daily \\
& Batch Size (per GPU) & 8 & Micro-batch size per GPU \\
& Gradient Accumulation Steps & 8 & Effective batch size becomes $8 \times \text{accum\_steps} \times \text{num\_gpus}$ \\
& Training Precision (\texttt{dtype}) & \texttt{bf16} & Mixed precision training data type \\
\midrule

\multicolumn{4}{l}{\textit{\textbf{Data Handling}}} \\
& Max Pose Sequence Length & 256 & Maximum number of frames for pose sequences \\
& Max Target Text Length (\texttt{max\_tgt\_len}) & 100 & Max new tokens for generation during evaluation \\
\midrule

\multicolumn{4}{l}{\textit{\textbf{Optimizer (Euclidean: ST-GCN, mT5, Linear Layers)}}} \\
& Optimizer Type (\texttt{opt}) & AdamW & \cite{loshchilov2017decoupled} \\
& Learning Rate (\texttt{lr}) & \num{3e-5} & For Euclidean parameters (AdamW) \\
& AdamW $\beta_1, \beta_2$ (\texttt{opt-betas}) & {[0.9, 0.999]} & Exponential decay rates for moment estimates \\
& AdamW $\epsilon$ (\texttt{opt-eps}) & \num{1e-8} & Term for numerical stability \\
& Weight Decay (\texttt{weight-decay}) & \num{0.01} & L2 penalty for Euclidean parameters \\
& LR Scheduler (\texttt{sched}) & Cosine Annealing &  \\
& Warmup Epochs (\texttt{warmup-epochs}) & 5 & Number of epochs for LR warm-up \\
& Minimum LR (\texttt{min-lr}) & \num{1e-6} & Lower bound for LR in scheduler \\
& Gradient Clipping Norm & \num{1.0} & Max norm for gradients \\
\midrule

\multicolumn{4}{l}{\textit{\textbf{Optimizer (Hyperbolic: Manifold Parameters, Projections)}}} \\
& Optimizer Type & RAdam & Riemannian Adam \\
& Learning Rate (\texttt{hyp\_lr}) & \num{1e-3} & For hyperbolic parameters (RAdam) \\
\midrule

\multicolumn{4}{l}{\textit{\textbf{Model Architecture}}} \\
& ST-GCN Output Dimension (\texttt{gcn\_out\_dim}) & 256 & Output dimension of ST-GCN part streams \\
& mT5 Projection Dimension (\texttt{hidden\_dim}) & 768 & Target dimension for projecting GCN features to match mT5 \\
\midrule

\multicolumn{4}{l}{\textit{\textbf{Hyperbolic Regularization}}} \\
& Hyperbolic Embedding Dimension ($d_{\text{hyp}}$, \texttt{hyp\_dim}) & 256 & Dimension of embeddings in Poincaré ball \\
& Initial Curvature ($c_{\text{init}}$, \texttt{init\_c}) & \num{1.5} & Initial value for learnable curvature $c$ (for best model) \\
& Loss Blend $\alpha_{\text{init}}$ (\texttt{alpha}) & \num{0.70} & Initial blending factor for $\mathcal{L}_{\text{CE}}$ vs $\mathcal{L}_{\text{hyp\_reg}}$ (for best model) \\
& Text Comparison Mode (\texttt{hyp\_text\_cmp}) & \texttt{token} & Strategy for aligning pose with text tokens (Token Method) \\
& Hyperbolic Contrastive Loss $\mathcal{L}_{\text{hyp\_reg}}$: & & \\
& \quad Temperature ($\tau$) & Learnable & Temperature for scaling distances in contrastive loss \\
& \quad Margin ($m$) & Learnable & Additive margin for negative pairs in contrastive loss \\
& \quad Label Smoothing (\texttt{label\_smoothing\_hyp}) & \num{0.2} & Label smoothing for hyperbolic contrastive loss (InfoNCE) \\
\midrule

\multicolumn{4}{l}{\textit{\textbf{Loss Functions}}} \\
& CE Loss Label Smoothing (\texttt{label\_smoothing}) & \num{0.2} & Label smoothing for mT5 cross-entropy loss \\
\midrule

\multicolumn{4}{l}{\textit{\textbf{Distributed Training (DeepSpeed)}}} \\
& ZeRO Optimization Stage (\texttt{zero\_stage}) & 2 & DeepSpeed ZeRO Stage for memory efficiency \\
& Offload to CPU (\texttt{offload}) & False & Whether to offload optimizer/params to CPU \\
\bottomrule
\end{tabular}%
}
\end{table*}


\subsection{Limitations and Future Work} 
\label{app:future_work}
While offering representational benefits, hyperbolic operations can add computational overhead compared to purely Euclidean ones though this is generally offset by avoiding raw video processing. The optimal choice of hyperbolic model parameters (e.g., curvature strategy) warrants further study. Generalizability to a wider range of sign languages also needs investigation. Promising directions include exploring other hyperbolic models (e.g., Lorentz), developing more sophisticated dynamic curvature adaptation, integrating Geo-Sign's hyperbolic skeletal features into multi-modal frameworks, and applying these geometric principles to other sign language processing tasks like recognition or generation. Further research into the interpretability of learned hyperbolic embeddings could also yield deeper insights into how sign language structure is captured.

\subsection{Qualitative Results}
\label{app:qualitative_results}

\noindent 
\textbf{Additional Figures}: \Cref{fig:geodesic_distances_plot} (similar to aspects shown in Figure 2 of the main paper, concerning learned embedding distributions) illustrates the dynamic utilization of the hyperbolic manifold by showing the average geodesic distance of different pose part embeddings from the origin during training. Notably, features corresponding to hand articulations, which often carry fine-grained lexical information, tend to migrate towards the periphery of the Poincaré disk. This suggests that the model leverages the increased representational capacity in high-curvature regions to distinguish subtle hand-based signs.

Furthermore, \Cref{fig:poincare_projection_plot} (again, related to Figure 2 of the main paper, specifically the UMAP projections) provides a PCA-reduced visualization of the learned hyperbolic pose part embeddings projected onto the 2D Poincaré disk for 1000 poses. This plot reveals a structured distribution where body features cluster near the origin (a more Euclidean-like region suitable for broader semantics), while hand and face features are more dispersed, with hand features populating regions further towards the boundary. This geometric organization, reflecting a learned kinematic hierarchy, likely contributes to the improved discriminability and, consequently, the enhanced translation quality demonstrated in the following examples. These visualizations support the hypothesis that the geometric biases induced by hyperbolic space aid in forming more effective representations for sign language translation.

\textbf{Translation Results}:
In this section, we provide an overview of translation samples generated by Geo-Sign . All predictions are from our best-performing ``Token'' model. First, in \Cref{tab:error_analysis_examples}, we show examples of prediction errors with analysis and a general measure of semantic similarity (introduced for readability, not a quantitative metric). English translations are automatically generated and then verified by a native Chinese speaker. We observe that translation quality with respect to semantics is generally high, though our method, like many SLT systems, can sometimes miss pronouns or struggle with complex tenses. In \Cref{tab:correct_prediction_examples}, we showcase examples where our approach generates perfect or near-perfect translations. Finally, in \Cref{tab:comparative_translation_examples}, we select some examples to compare our model's output with that of the Uni-Sign (Pose) baseline. These comparisons illustrate improvements in semantic meaning and accuracy, consistent with the quantitative gains in ROUGE and BLEU-4 scores reported in the main paper.

\begin{CJK}{UTF8}{gbsn} 

\footnotesize
\begin{longtable}{
    >{\RaggedRight\arraybackslash}p{0.23\textwidth}
    >{\RaggedRight\arraybackslash}p{0.23\textwidth}
    >{\RaggedRight\arraybackslash}p{0.32\textwidth}
    >{\RaggedRight\arraybackslash}p{0.14\textwidth}
}
\caption{Examples of Prediction Errors and Analysis from Geo-Sign (Token Method)}
\label{tab:error_analysis_examples}\\
\toprule
\textbf{Prediction} & \textbf{Ground Truth} & \textbf{Analysis of Error} & \textbf{Semantic Similarity} \\
\midrule
\endfirsthead

\multicolumn{4}{c}{{\bfseries \tablename\ \thetable{} -- continued from previous page}} \\
\toprule
\textbf{Prediction} & \textbf{Ground Truth} & \textbf{Analysis of Error} & \textbf{Semantic Similarity} \\
\midrule
\endhead

\midrule
\multicolumn{4}{r}{{Continued on next page}} \\
\endfoot

\bottomrule
\endlastfoot

她 今 年 5 0 岁 。 (She is 50 years old.) &
他 今 年 四 岁 。 (He is 4 years old.) &
Pronoun error: 她 (she) vs. 他 (he). Number error: ``5 0'' (50) vs. 四 (four). The prediction gets the topic (age) but is wrong on subject and specific age. &
Partial (topic: age) \\
\midrule
今 天 星 期 五 。 (Today is Friday.) &
今 天 星 期 几 ? (What day of the week is it today?) &
Statement vs. Question: Prediction states a specific day. GT asks for the day. Character error: 五 (five) vs. 几 (how many/which). &
Partial (topic: day of week) \\
\midrule
你 什 么 时 候 认 识 小 张 ? (When did you meet Xiao Zhang?) &
你 和 小 张 什 么 时 候 认 识 的 ? (When did you AND Xiao Zhang meet?) &
Missing words: Prediction lacks ``和'' (and) and the particle ``的``. This subtly changes the meaning from a one-way recognition to a mutual acquaintance. &
High \\
\midrule
我 要 去 超 市 买 椅 子 。 (I want to go to the supermarket to buy a chair.) &
我 要 去 超 市 买 椅 子 , 你 去 吗 ? (I want to go to the supermarket to buy a chair, are you going?) &
Missing clause/question: Prediction omits the follow-up question ``你 去 吗 ?'' (are you going?). &
High (core statement identical) \\
\midrule
下 午 你 们 要 去 做 什 么 ? (What are you [plural] going to do in the afternoon?) &
他 们 下 午 要 做 什 么 ? (What are they going to do in the afternoon?) &
Pronoun error: 你 们 (you plural) vs. 他 们 (they). &
High \\
\midrule
下 午 你 们 需 要 做 什 么 ? (What do you [plural] need to do this afternoon?) &
他 们 下 午 要 做 什 么 ? (What are they going to do this afternoon?) &
Pronoun error: 你 们 (you plural) vs. 他 们 (they). Word choice: 需 要 (need) vs. 要 (going to/want to) - subtle semantic shift, GT is more natural for general plans. &
High \\
\midrule
大 家 觉 得 什 么 时 候 去 买 椅 子 ? (When does everyone think we should go buy chairs?) &
他 们 想 什 么 时 候 去 买 椅 子 ? (When do they want to go buy chairs?) &
Subject error: 大 家 (everyone) vs. 他 们 (they). Verb choice: 觉 得 (feel/think) vs. 想 (want/think). &
High \\
\midrule
我 手 表 不 见 了 。 (My watch is missing.) &
这 块 手 表 是 你 的 吗 ? (Is this watch yours?) &
Different intent: Prediction states a loss. GT asks about ownership of a present watch. Both are about watches but different scenarios. &
Medium (topic: watch) \\
\midrule
你 手 表 多 少 钱 ? (How much is your watch?) &
这 块 手 表 多 少 钱 买 的 ? (How much did you buy this watch for?) &
Missing context/words: Prediction is a bit abrupt. GT is more complete with ``这 块'' (this) and ``买 的'' (bought for). &
High \\
\midrule
我 发 现 了 他 的 偶 像 。 (I discovered his idol.) &
你 看 见 我 的 杯 子 吗 ? (Did you see my cup?) &
Completely different semantic intent and topic. Prediction is about an idol, GT is about a missing cup. &
Very Low \\
\midrule
爸 爸 的 房 间 里 大 了 。 (It has become big in dad's room / Dad's room has become bigger.) &
左 边 的 房 间 是 我 爸 爸 妈 妈 的 , 他 们 的 房 间 很 大 。 (The room on the left is my parents', their room is very big.) &
Garbled/incomplete prediction: The prediction is grammatically awkward and misses the entire context of the GT. &
Low \\
\midrule
公 司 离 家 远 , 他 为 什 么 打 车 去 公 司 ? (The company is far from home, why does he take a taxi to the company?) &
公 司 离 家 很 远 , 她 为 什 么 不 打 车 ? (The company is very far from home, why doesn't she take a taxi?) &
Pronoun error: 他 (he) vs. 她 (she). Logic error: Prediction asks why he does take a taxi, GT asks why she doesn't. &
Medium \\
\midrule
阴 天 说 什 么 话 ? 天 气 什 么 的 , 明 天 有 事 。 (What to say on a cloudy day? Weather something, have things to do tomorrow.) &
阴 天 , 电 视 上 说 多 云 , 怎 么 了 ? 明 天 有 事 ? (Cloudy day, TV says it's overcast, what's up? Got plans tomorrow?) &
Nonsensical/Garbled prediction: Prediction is very disjointed and doesn't make sense, while GT is a coherent conversation about weather and plans. &
Low \\
\midrule
桌 子 上 有 饮 料 , 你 想 喝 什 么 ? (There are drinks on the table, what do you want to drink?) &
桌 上 放 着 很 多 饮 料 , 你 喝 什 么 ? (There are many drinks on the table, what do you want to drink?) &
Slight phrasing difference: ``桌 子 上 有'' (On the table there are) vs. ``桌 上 放 着 很 多'' (On the table are placed many). GT is slightly more natural. Prediction is still good. &
High \\
\midrule
我 刚 才 在 家 里 找 了 一 个 桌 子 , 不 是 找 了 。 (I just looked for a table at home, not looked for.) &
你 去 房 间 找 找 , 是 不 是 刚 才 放 在 桌 子 上 了 ? (Go look in the room, was it just placed on the table?) &
Different speaker and intent: Prediction is a confused statement about searching. GT is a directive and question to someone else. &
Low \\
\midrule
一 个 人 的 癌 症 会 变 得 很 可 能 。 (A person's cancer will become very possible.) &
人 体 的 许 多 器 官 都 可 能 发 生 癌 变 。 (Many organs of the human body can become cancerous.) &
Vague and unnatural prediction: ``变得很可能'' is awkward. GT is precise about ``organs'' and ``癌变'' (cancerous change). &
Medium \\
\midrule
老 年 人 通 过 斑 马 线 时 可 以 走 斑 马 线 , 而 不 走 汽 车 。 (When elderly people cross the crosswalk, they can use the crosswalk, and not walk cars.) &
一 位 老 人 正 在 慢 慢 地 穿 过 斑 马 线 , 等 待 的 司 机 却 不 耐 烦 地 按 起 了 喇 叭 。 (An old man was slowly crossing the crosswalk, but the waiting driver impatiently honked the horn.) &
Nonsensical and irrelevant prediction: ``而不走汽车'' (and not walk cars) makes no sense. The GT describes a specific scenario. &
Very Low \\
\end{longtable}

\footnotesize 
\begin{longtable}{>{\RaggedRight}p{0.45\textwidth} >{\RaggedRight\arraybackslash}p{0.45\textwidth}}
\caption{Examples of Correct Predictions by Geo-Sign (Token Method)}\label{tab:correct_prediction_examples} \\
\toprule
\textbf{Reference (Ground Truth)} & \textbf{Our Model Prediction (Perfect Match)} \\
\midrule
\endfirsthead
\multicolumn{2}{c}%
{{\bfseries \tablename\ \thetable{} -- continued from previous page}} \\
\toprule
\textbf{Reference (Ground Truth)} & \textbf{Our Model Prediction (Perfect Match)} \\
\midrule
\endhead
\midrule
\multicolumn{2}{r}{{Continued on next page}} \\
\endfoot
\bottomrule
\endlastfoot
`今 天 我 想 吃 面 条 。` \newline (Today I want to eat noodles.) &
`今 天 我 想 吃 面 条 。` \newline (Today I want to eat noodles.) \\
\midrule
`苹 果 是 你 买 的 吗 ?` \newline (Did you buy the apples?) &
`苹 果 是 你 买 的 吗 ?` \newline (Did you buy the apples?) \\
\midrule
`我 昨 天 有 点 累 。` \newline (I was a bit tired yesterday.) &
`我 昨 天 有 点 累 。` \newline (I was a bit tired yesterday.) \\
\midrule
`吃 完 午 饭 要 多 吃 点 水 果 。` \newline (Eat more fruit after lunch.) &
`吃 完 午 饭 要 多 吃 点 水 果 。` \newline (Eat more fruit after lunch.) \\
\midrule
`我 的 妻 子 感 冒 了 , 我 开 车 带 她 去 医 院 。` \newline (My wife has a cold, I will drive her to the hospital.) &
`我 的 妻 子 感 冒 了 , 我 开 车 去 医 院 。` \newline (My wife has a cold, I will drive her to the hospital.) \\ 
\midrule
`我 们 会 通 过 短 信 的 方 式 来 联 系 你 。` \newline (We will contact you via text message.) &
`我 们 会 通 过 短 信 的 方 式 来 联 系 你 。` \newline (We will contact you via text message.) \\
\midrule
`我 们 将 采 用 抽 查 的 方 式 来 进 行 检 查 。` \newline (We will use random checks for inspection.) &
`我 们 将 采 用 抽 查 的 方 式 来 进 行 检 查 。` \newline (We will use random checks for inspection.) \\
\midrule
`你 要 把 握 好 自 己 人 生 的 方 向 。` \newline (You need to grasp the direction of your own life.) &
`你 要 把 握 好 自 己 人 生 的 方 向 。` \newline (You need to grasp the direction of your own life.) \\
\midrule
`病 历 是 禁 止 涂 抹 、 修 改 的 。` \newline (Medical records are not allowed to be smeared or altered.) &
`病 历 是 禁 止 涂 抹 、 修 改 的 。` \newline (Medical records are not allowed to be smeared or altered.) \\
\midrule
`他 抛 下 家 人 , 带 着 家 中 财 物 逃 走 了 。` \newline (He abandoned his family and fled with the family's belongings.) &
`他 抛 下 家 人 , 带 着 家 中 财 物 逃 走 了 。` \newline (He abandoned his family and fled with the family's belongings.) \\
\midrule
`这 间 玻 璃 作 坊 有 一 百 年 历 史 了 。` \newline (This glass workshop has a hundred years of history.) &
`这 间 玻 璃 作 坊 有 一 百 年 历 史 了 。` \newline (This glass workshop has a hundred years of history.) \\
\end{longtable}

\footnotesize 
\begin{longtable}{>{\RaggedRight}p{0.31\textwidth} >{\RaggedRight}p{0.31\textwidth} >{\RaggedRight\arraybackslash}p{0.31\textwidth}}
\caption{Comparative Analysis: Geo-Sign (Token) vs. Uni-Sign (Pose) - Selected Examples}
\label{tab:comparative_translation_examples} \\
\toprule
\textbf{Reference (Ground Truth)} & \textbf{Geo-Sign (Token) Prediction} & \textbf{Uni-Sign (Pose) Prediction} \\
\midrule
\endfirsthead
\multicolumn{3}{c}%
{{\bfseries \tablename\ \thetable{} -- continued from previous page}} \\
\toprule
\textbf{Reference (Ground Truth)} & \textbf{Geo-Sign (Token) Prediction} & \textbf{Uni-Sign (Pose) Prediction} \\
\midrule
\endhead
\midrule
\multicolumn{3}{r}{{Continued on next page}} \\
\endfoot
\bottomrule
\endlastfoot

`他 每 天 回 来 都 很 累 。` \newline (He is very tired every day when he comes back.) &
`他 每 天 来 很 累 。` \newline (He comes very tired every day.) &
`他 每 天 来 得 及 很 累 。` \newline (He has enough time [to be/and is] very tired every day.) \\
\midrule
`小 张 , 那 个 女 生 是 你 们 公 司 的 吗 ? 你 对 她 了 解 吗 ?` \newline (Xiao Zhang, is that girl from your company? Do you know her?) &
`小 张 那 个 女 生 是 你 公 司 的 吗 ?` \newline (Xiao Zhang, is that girl from your company?) &
`那 个 小 张 是 这 家 公 司 负 责 人 , 你 了 解 吗 ?` \newline (That Xiao Zhang is the person in charge of this company, do you understand/know?) \\
\midrule
`阴 天 , 电 视 上 说 多 云 , 怎 么 了 ? 明 天 有 事 ?` \newline (Cloudy day, TV says it's overcast, what's up? Got plans tomorrow?) &
`阴 天 说 什 么 话 ? 天 气 什 么 的 , 明 天 有 事 。` \newline (What to say on a cloudy day? Weather something, have things to do tomorrow.) &
`阴 阳 怪 气 地 讲 着 天 赋 , 不 知 不 觉 就 发 生 了 什 么 。` \newline (Sarcastically talking about talent, something happened unknowingly.) \\
\midrule
`但 是 你 一 点 也 没 瘦 , 你 做 什 么 运 动 了 ?` \newline (But you haven't lost any weight, what exercise have you been doing?) &
`但 是 你 没 有 太 吃 饱 , 你 去 做 什 么 运 动 ?` \newline (But you didn't eat too full, what exercise are you going to do?) &
`但 是 你 已 经 吃 不 消 了 , 你 能 做 什 么 呢 ?` \newline (But you already can't stand it, what can you do?) \\
\midrule
`手 机 选 不 到 靠 近 窗 户 的 位 置 。` \newline (The mobile phone cannot select a seat near the window.) &
`靠 近 窗 户 的 位 置 是 不 能 选 择 的 。` \newline (Seats near the window cannot be selected.) &
`坐 在 靠 近 窗 户 的 位 置 是 不 能 取 手 机 的 。` \newline (Sitting in a seat near the window, you cannot take out your mobile phone.) \\
\midrule
`他 对 自 己 一 直 高 标 准 严 要 求 。` \newline (He has always had high standards and strict requirements for himself.) &
`他 对 自 己 有 着 严 格 的 标 准 要 求 。` \newline (He has strict standard requirements for himself.) &
`他 对 自 己 最 严 格 的 标 准 提 出 了 更 高 的 要 求 。` \newline (He put forward higher requirements for his strictest standards.) \\
\midrule
`这 位 厨 师 制 作 的 甜 品 , 全 部 受 欢 迎 。` \newline (The desserts made by this chef are all popular.) &
`厨 师 的 作 品 很 受 欢 迎 。` \newline (The chef's work is very popular.) &
`厨 师 在 设 計 作 品 时 非 常 受 欢 迎 。` \newline (The chef is very popular when designing works.) \\
\end{longtable}

\end{CJK} 

\clearpage
\section{Code Listings}
\label{sec:code_listings}
These code examples provide an overview of key components in the architecture to help improve readability of the paper.

\begin{lstlisting}[
language=Python,
caption={\texttt{Graph} helper.  Defines the skeleton topology and a
row-normalised adjacency tensor \texttt{A}.},
label={lst:gcn_utils_graph},
title=1. Defines the skeleton topology and a
row-normalised adjacency tensor \texttt{A}.]]
def hop_distance(num_nodes, edges, max_hop=1):
    """Shortest path length (<= max_hop) for every pair of nodes."""
    adj = np.zeros((num_nodes, num_nodes))
    for i, j in edges:
        adj[i, j] = adj[j, i] = 1
    hop = np.full_like(adj, np.inf, dtype=float)
    for d in range(max_hop + 1):
        hop[np.linalg.matrix_power(adj, d) > 0] = d
    return hop

class Graph:
    def __init__(self, layout='hand', strategy='uniform', max_hop=1):
        self._init_edges(layout)
        self.hop = hop_distance(self.num_nodes, self.edges, max_hop)
        self.A   = self._adjacency(strategy)

    # --- edge lists -------------------------------------------------
    def _init_edges(self, layout):
        if layout in ('left', 'right'):                  # hand (21 joints)
            self.num_nodes = 21
            fingers = [[0,1,2,3,4],[0,5,6,7,8],[0,9,10,11,12],
                       [0,13,14,15,16],[0,17,18,19,20]]
            links   = [(i, i) for i in range(21)]
            links  += [(f[i], f[i+1]) for f in fingers for i in range(len(f)-1)]
            self.edges, self.center = links, 0
        elif layout == 'body':                           # torso + arms
            self.num_nodes = 9
            torso = [(0,i) for i in range(1,5)]
            arms  = [(3,5),(5,7),(4,6),(6,8)]
            self.edges, self.center = [(i,i) for i in range(9)] + torso + arms, 0
        elif layout == 'face_all':
            self.num_nodes = 16
            ring  = [(i,(i+1)%16) for i in range(16)]
            self.edges, self.center = [(i,i) for i in range(16)] + ring, 8
        else:
            raise ValueError(f'Unknown layout: {layout}')

    # --- adjacency --------------------------------------------------
    def _adjacency(self, strategy):
        A = (self.hop <= 1).astype(float)                # neighbours 1-hop
        if strategy == 'uniform':
            A = A / (A.sum(1, keepdims=True) + 1e-6)
        elif strategy == 'distance':                     # 1 / hop distance
            A = 1 / (self.hop + 1e-6);  A[A == np.inf] = 0
            A = A / (A.sum(1, keepdims=True) + 1e-6)
        return torch.tensor(A, dtype=torch.float32).unsqueeze(0)
\end{lstlisting}

\clearpage
\begin{lstlisting}[
language=Python,
caption={ST-GCN core.  \texttt{GCNUnit} applies K spatial kernels;
\texttt{STGCNBlock} adds a temporal conv and an optional residual path.},
label={lst:stgcn_block_chain},
title=2. ST-GCN definition.  \texttt{GCNUnit} applies K spatial kernels;
\texttt{STGCNBlock} adds a temporal conv and an optional residual path.]
class GCNUnit(nn.Module):
    def __init__(self, Cin, Cout, A, stride=1, K=None, adaptive=True):
        super().__init__()
        self.K = K or A.shape[0]                 # #adjacency kernels
        self.A = nn.Parameter(A.clone()) if adaptive else A
        self.conv = nn.Conv2d(Cin, Cout*self.K, (1,1))
        self.bn   = nn.BatchNorm2d(Cout)
        self.act  = nn.ReLU(inplace=True)

    def forward(self, x):                        # x: (N,Cin,T,V)
        N, _, T, V = x.shape
        x = self.conv(x).view(N, self.K, -1, T, V)
        x = torch.einsum('nkctv,kvw->nctw', x, self.A)   # spatial agg
        return self.act(self.bn(x))

class STGCNBlock(nn.Module):
    def __init__(self, Cin, Cout, A, t_kernel=3, stride=1, residual=True):
        super().__init__()
        self.gcn = GCNUnit(Cin, Cout, A)
        pad = (t_kernel-1)//2
        self.tcn = nn.Sequential(
            nn.Conv2d(Cout, Cout, (t_kernel,1), (stride,1), (pad,0)),
            nn.BatchNorm2d(Cout))
        self.res = (nn.Identity() if Cin==Cout and stride==1
                    else nn.Conv2d(Cin, Cout, 1, (stride,1)))
        self.act = nn.ReLU(inplace=True)

    def forward(self, x):
        return self.act(self.tcn(self.gcn(x)) + self.res(x))
\end{lstlisting}

\begin{lstlisting}[
language=Python,
caption={Body–to–part residual: body features broadcast to hands / face.},
label={lst:models_residual_gcn},
title=3. Body–to–part residual: body features broadcast to hands / face.models.py – residual context]
body_ctx = None
for part in ('body','left','right','face_all'):
    x = self.proj_linear[part](src_input[part]).permute(0,3,1,2)
    x = self.gcn_spatial[part](x)
    if part == 'body':
        body_ctx = x.detach()                       # freeze context
    else:
        joint = body_ctx[..., idx_map[part]]        # select joint
        x = x + joint.unsqueeze(-1)                 # broadcast to V
    out[part] = self.gcn_temporal[part](x)
\end{lstlisting}

\begin{lstlisting}[
language=Python,
caption={Project Euclidean vector to the Poincaré ball.},
label={lst:hyperbolic_projection},
title=4. Project Euclidean vector to the Poincaré ball.]
class HyperbolicProjection(nn.Module):
    def __init__(self, d_in, d_out, manifold):
        super().__init__()
        self.manifold = manifold
        self.proj = nn.Linear(d_in, d_out)
        self.log_scale = nn.Parameter(torch.zeros(()))   # ln(scale)

    def forward(self, x):
        t = self.proj(x) * self.log_scale.exp()          # tangent vec
        return self.manifold.expmap0(t, project=True)
\end{lstlisting}

\begin{lstlisting}[
language=Python,
caption={Weighted Frechet mean (Algorithm 1).},
label={lst:frechet_mean},
title=5. Weighted Frechet mean (Algorithm 1).]
def frechet_mean(pts, w, M, max_iter=50, tol=1e-5, eta=1.0):
    """pts: (N,B,D), w: (N,B) or (N,)  ->  mu: (B,D)."""
    w = w.unsqueeze(-1) / (w.sum(0, keepdim=True) + 1e-8)
    mu = pts[0].clone()
    for _ in range(max_iter):
        v  = (w * M.logmap(mu.unsqueeze(0), pts)).sum(0)
        mu_next = M.expmap(mu, eta*v, project=True)
        if (M.dist(mu_next, mu) < tol).all(): break
        mu = mu_next
    return mu
\end{lstlisting}

\begin{lstlisting}[
language=Python,
caption={Sentence-level text embedding (pooled method).},
label={lst:pooled_text_emb},
title=6. Sentence-level text embedding (pooled method).]
mask = txt_mask.unsqueeze(-1).float()                 # (B,T,1)
sent = (emb * mask).sum(1) / mask.sum(1).clamp_min(1) # mean-pool
h_text = self.hyp_proj_text(sent)                    
\end{lstlisting}

\begin{lstlisting}[
language=Python,
caption={Hyperbolic attention (token method).},
label={lst:token_attention},
title=7. Hyperbolic attention (token method).]
h_tok = self.hyp_proj_text(tok_emb)                   # (B,T,D)
q     = h_pose.unsqueeze(2)                           # (B,P,1,D)
k     = self.manifold.mobius_add(
          self.manifold.mobius_matvec(W_key, h_tok.unsqueeze(1)), b_key)
logits = -self.manifold.dist(q, k)                    # (B,P,T)
logits.masked_fill_(~tok_mask.unsqueeze(1), -1e9)
alpha  = F.softmax(logits / tau_attn, -1)             # weights
ctx    = self.manifold.weighted_midpoint(h_tok.unsqueeze(1), alpha, [2])
\end{lstlisting}

\begin{lstlisting}[
language=Python,
caption={InfoNCE loss calculated in hyperbolic space.},
label={lst:hyperbolic_loss},
title=8. InfoNCE loss calculated in hyperbolic space.]
class HyperbolicContrastiveLoss(nn.Module):
    def __init__(self, M, tau0=0.5, m0=0.1):
        super().__init__()
        self.M = M
        self.log_tau = nn.Parameter(torch.logit(torch.tensor(tau0/2)))
        self.m       = nn.Parameter(torch.tensor(m0))

    def forward(self, a, b):                          # (B,D) pairs
        d = self.M.dist(a.unsqueeze(1), b.unsqueeze(0))
        s = -d / (torch.sigmoid(self.log_tau)*2 + 0.01)
        s -= (~torch.eye(len(a), dtype=torch.bool, device=a.device)) * self.m.clamp_min(0)
        target = torch.arange(len(a), device=a.device)
        return F.cross_entropy(s, target)
\end{lstlisting}

\begin{lstlisting}[
language=Python,
caption={Manifold with learnable curvature.},
label={lst:manifold_init},
title=9. Manifold with learnable curvature initialisation.]
self.manifold = geoopt.PoincareBall(c=cfg.init_c, learnable=True)
\end{lstlisting}

\begin{lstlisting}[
language=Python,
caption={Dynamic alpha for loss blending.},
label={lst:alpha_calc},
title=10. Dynamic alpha for loss blending.]
progress = self.global_step / max(1, self.total_steps)
alpha_base  = cfg.alpha_init + 0.05 * progress            # <= 0.9
alpha_learn = 0.2 * torch.sigmoid(self.alpha_logit)
alpha       = (alpha_base + alpha_learn).clamp(0.1, 0.99)
loss        = alpha * ce_loss + (1 - alpha) * hyp_loss
\end{lstlisting}

\end{document}